\DocumentMetadata{
  pdfversion=2.0,pdfstandard=ua-2,
}

\documentclass[sigconf,screen]{acmart-tagged}  
\AtBeginDocument{%
  }

\newif\ifincludeappendix
\includeappendixtrue  

\newcommand{\appref}[1]{%
  \ifincludeappendix
    ~\ref{#1}%
  \else
    ~\cite{appendix}%
  \fi
}

\setcopyright{cc}
\setcctype{by-nc-nd}
\acmDOI{10.1145/3757279.3788668}
\acmYear{2026}
\copyrightyear{2026}
\acmISBN{979-8-4007-2128-1/2026/03}
\acmConference[HRI '26]{Proceedings of the 21st ACM/IEEE International Conference on Human-Robot Interaction}{March 16--19, 2026}{Edinburgh, Scotland, UK}
\acmBooktitle{Proceedings of the 21st ACM/IEEE International Conference on Human-Robot Interaction (HRI '26), March 16--19, 2026, Edinburgh, Scotland, UK}
\received{2025-09-30}
\received[accepted]{2025-12-23}

\usepackage{algpseudocode}              
\usepackage{algorithm}                  
\usepackage{bbm}
\usepackage{makecell}                   
\usepackage{subcaption}                 
\usepackage{placeins}
\usepackage{enumitem}
\usepackage[none]{hyphenat}
\usepackage{ragged2e}

\makeatletter
\renewcommand{\fnum@algorithm}{\algorithmname\space\thealgorithm}
\makeatother

\newcommand{\affiliationtight}[1]{%
  \affiliation{\vspace{-1.5pt}#1\vspace{-3.5pt}}
}
\newcommand{\fnref}[1]{\hyperref[#1]{\textsuperscript{\ref*{#1}}}}

\begin{document}

\title{A Human-in-the-Loop Confidence-Aware Failure Recovery Framework for Modular Robot Policies}




\author{Rohan Banerjee}
\affiliationtight{%
  \institution{Cornell University}
 \country{USA}
}

\author{Krishna Palempalli}
\authornote{Equal Contribution.}
\affiliationtight{%
  \institution{Cornell University}
   \country{USA}
}

\author{Bohan Yang}
\authornotemark[1]
\affiliationtight{%
  \institution{Cornell University}
   \country{USA}
}

\author{Jiaying Fang}
\affiliationtight{%
  \institution{Cornell University}
   \country{USA}
}

\author{Alif Abdullah}
\affiliationtight{%
  \institution{Cornell University}
 \country{USA}
}

\author{Tom Silver}
\affiliationtight{%
  \institution{Princeton University}
 \country{USA}
}

\author{Sarah Dean}
\authornote{Equal Advising.}
\affiliationtight{%
  \institution{Cornell University}
 \country{USA}
}

\author{Tapomayukh Bhattacharjee}
\authornotemark[2]
\affiliationtight{%
  \institution{Cornell University}
 \country{USA}
}

\renewcommand{\shortauthors}{Rohan Banerjee et al.}

\renewcommand{\topfraction}{0.95}
\renewcommand{\dbltopfraction}{0.95}
\renewcommand{\textfraction}{0.05}
\renewcommand{\floatpagefraction}{0.9}


\begin{abstract}
\vspace{-0.1cm}

Robots operating in unstructured human environments inevitably encounter failures, especially in robot caregiving scenarios. While humans can often help robots recover, excessive or poorly targeted queries impose unnecessary cognitive and physical workload on the human partner. We present a human-in-the-loop failure-recovery framework for modular robotic policies, where a policy is composed of distinct modules such as perception, planning, and control, any of which may fail and often require different forms of human feedback. Our framework integrates calibrated estimates of module-level uncertainty with models of human intervention cost to decide which module to query and when to query the human. It separates these two decisions: a module selector identifies the module most likely responsible for failure, and a querying algorithm determines whether to solicit human input or act autonomously. We evaluate several module-selection strategies and querying algorithms in controlled synthetic experiments, revealing trade-offs between recovery efficiency, robustness to system and user variables, and user workload. Finally, we deploy the framework on a robot-assisted bite acquisition system and demonstrate, in studies involving individuals with both emulated and real mobility limitations, that it improves recovery success while reducing the workload imposed on users. Our results highlight how explicitly reasoning about both robot uncertainty and human effort can enable more efficient and user-centered failure recovery in collaborative robots. Supplementary materials and videos can be found at: \href{http://emprise.cs.cornell.edu/modularhil}{\textbf{emprise.cs.cornell.edu/modularhil}}.

\end{abstract}




\begin{teaserfigure}
  \vspace{-0.3cm}
  \centering
  \includegraphics[width=0.75\textwidth,alt={Composite figure illustrating a human-in-the-loop robot-assisted feeding framework evaluated with in-lab and in-home studies. Left panels show participants with emulated and real mobility limitations interacting with a feeding robot. The right panels depict the modular failure recovery pipeline for an example of acquiring green beans, where failures occur due to incorrect skill parameter selection. A human-in-the-loop intervention queries the user based on confidence and query cost, incorporates user feedback, and successfully recovers from failure on a subsequent execution attempt.}]{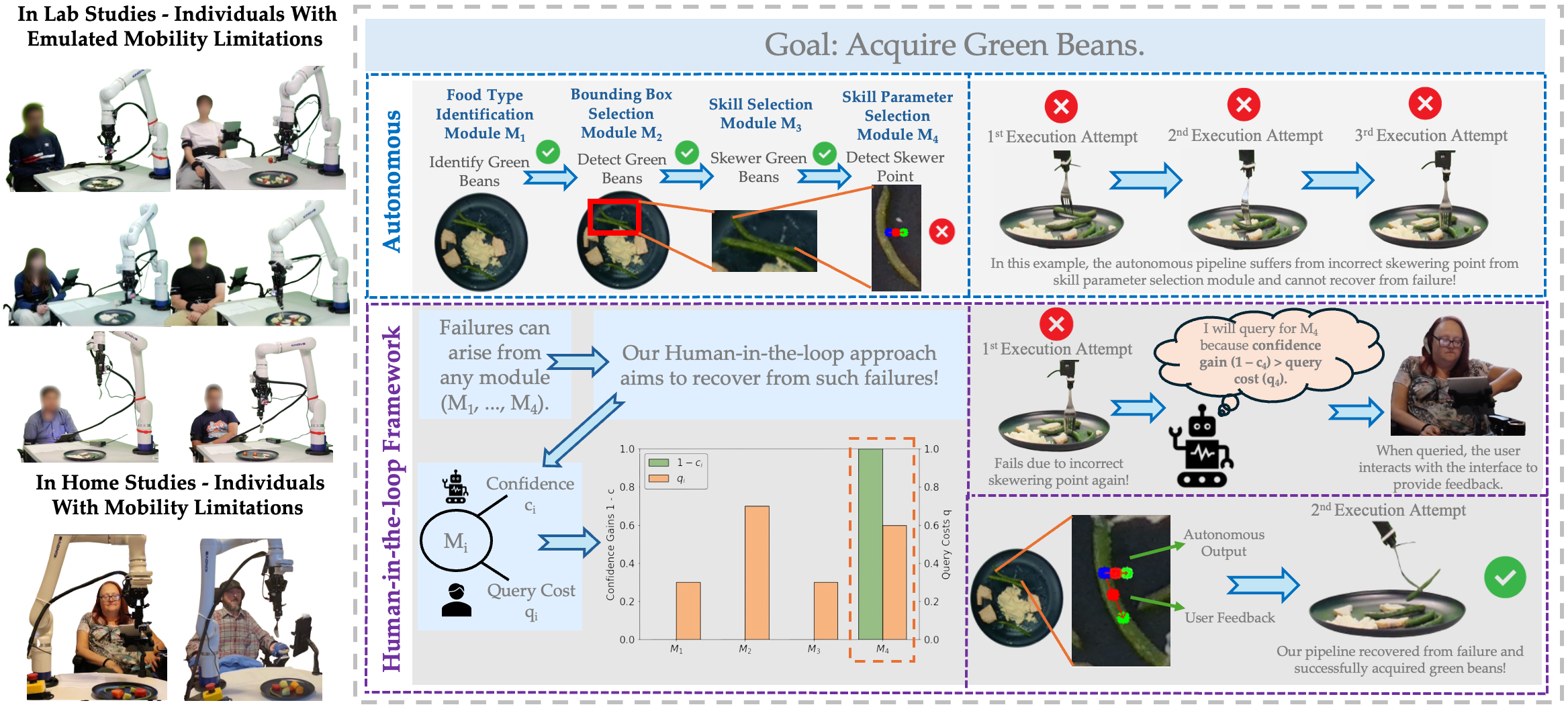}
     \vspace{-0.25cm}
  \caption{Our human-in-the-loop framework for failure recovery leverages confidence estimates from a modular policy, along with predicted estimates of user workload, to decide what to ask and when to act autonomously.}
  \Description{Composite figure illustrating a human-in-the-loop robot-assisted feeding framework evaluated with in-lab and in-home studies. Left panels show participants with emulated and real mobility limitations interacting with a feeding robot. The right panels depict the modular failure recovery pipeline for an example of acquiring green beans, where failures occur due to incorrect skill parameter selection. A human-in-the-loop intervention queries the user based on confidence and query cost, incorporates user feedback, and successfully recovers from failure on a subsequent execution attempt.}
  \label{fig:teaser}
\end{teaserfigure}


\maketitle


\begin{figure*}[t]
  \centering
  \includegraphics[width=0.75\textwidth, alt={Diagram of a human-in-the-loop bite acquisition system with a query policy. An RGB image containing food items is processed by a modular architecture consisting of four modules: food type identification, bounding box selection, skill selection, and skill parameter selection. The module confidences inform a query policy that decides whether to execute autonomously or query the user via a voice or tablet interface (consisting of both a module selector component, and a querying algorithm component). If execution fails, the system re-queries and retries until success.}]{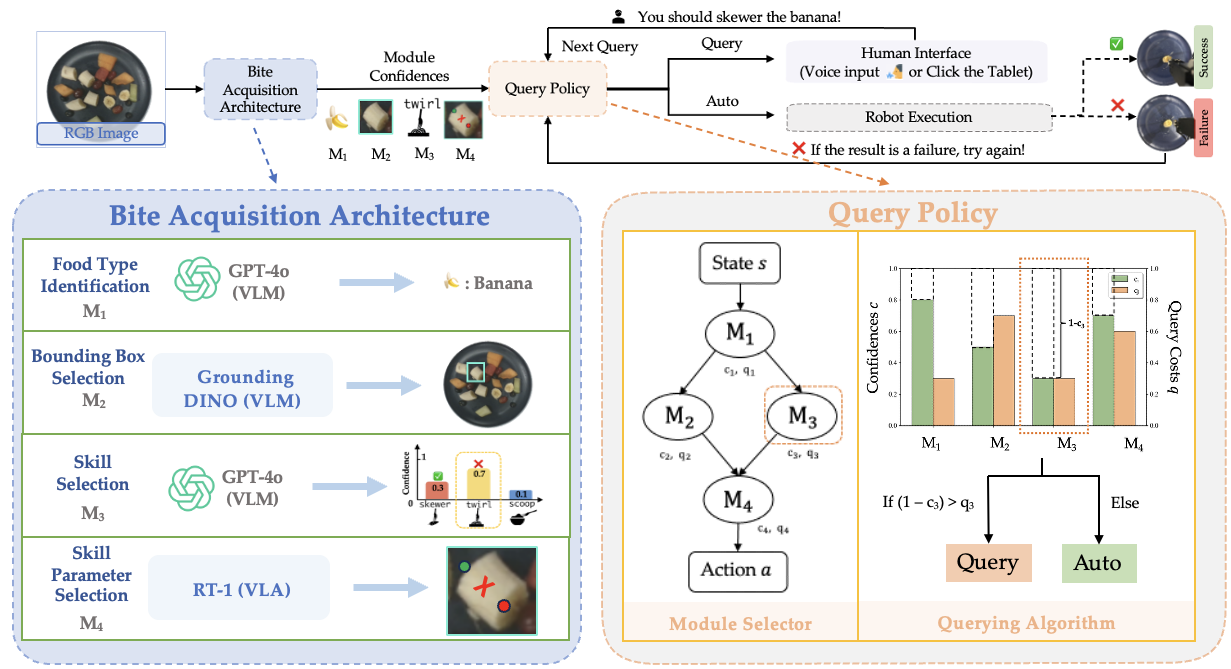}
   \vspace{-0.25cm}
  \caption{Overall human-in-the-loop decision failure recovery framework, grounded in the robot-assisted bite acquisition domain. The recovery framework first calls a module selector to decide which of the modules to query for (e.g. the skill selector). The framework then calls a querying algorithm, which decides whether to ask the user for help or act autonomously.}
  \Description{Diagram of a human-in-the-loop bite acquisition system with a query policy. An RGB image containing food items is processed by a modular architecture consisting of four modules: food type identification, bounding box selection, skill selection, and skill parameter selection. The module confidences inform a query policy that decides whether to execute autonomously or query the user via a voice or tablet interface (consisting of both a module selector component, and a querying algorithm component). If execution fails, the system re-queries and retries until success.}
  \label{fig:method}
\end{figure*}

\vspace{-2ex}
\section{Introduction}

Robots deployed in the wild inevitably fail, especially when assisting individuals with mobility limitations in performing activities of daily living (ADLs) in unstructured home environments~\cite{wu2025savor, gordon2020adaptive, sundaresan2022learning}. While humans can help robots recover~\cite{banerjee2024ask, fitzgerald2022inquire}, excessive querying imposes cognitive and physical workload on the human partner~\cite{hart1988development,smith2022decomposed}. Designing robots that know what and when to ask humans is therefore central to effective human–robot interaction. We focus on modular robot systems, which are more interpretable than end-to-end vision–language–action (VLA) policies, and thus more amenable to structured failure recovery. However, failures can arise in any of the perception, planning, or control modules, each requiring different forms of human feedback.

Failure recovery in modular systems is difficult for two reasons. First, identifying which module has failed is non-trivial; a perception error can cascade into planning and execution, making it unclear where intervention is most effective. Second, module confidence scores are often miscalibrated and do not reliably predict whether a task will succeed. Over-querying risks frustrating users, while under-querying risks repeated failures that undermine both task performance and trust. Properly balancing task success and human workload\footnote{In this work, we use the terms \textit{workload} and \textit{query cost} interchangeably.} lies at the heart of collaborative autonomy. By combining \emph{confidence-aware reasoning} with models of human workload~\cite{fridman2018cognitive, rajavenkatanarayanan2020towards, banerjee2024ask}, we can design \emph{workload-aware interaction} strategies that recover from failures more efficiently with high user satisfaction. This is crucial in assistive settings such as physical robot caregiving, where reliable task performance and user experience affect the acceptability of robot assistance.

To address the challenge of balancing recovery efficiency with user satisfaction, we propose a human-in-the-loop failure recovery framework for modular robot policies that integrates calibrated module-level uncertainty with models of human workload. The framework makes two key decisions at every recovery attempt. First, a \emph{module selector} determines which component of the modular policy to query, e.g. perception or planning. Second, a \emph{querying algorithm} determines whether to query the human at all or to act autonomously. This decision is critical for trading off the expected gain in task-level success against user workload. We formulate these decisions within a sequential decision-making framework, and evaluate several module-selection strategies and querying algorithms. We then deploy the framework on a robot-assisted bite acquisition system~\cite{gordon2020adaptive, sundaresan2022learning, gordon2021leveraging, banerjee2024ask}, demonstrating improved recovery success and reduced user querying workload across studies involving individuals with both emulated and real mobility limitations. 
Our work makes the following novel contributions:
\vspace{-1ex}
{
\tolerance=1000
\emergencystretch=0em
\begin{itemize}[leftmargin=*]
    \item We propose a modular human-in-the-loop framework for failure recovery that makes two complementary decisions: a \emph{module selector} chooses a policy component to query, and a \emph{querying algorithm} decides to solicit human input or act autonomously.
    \item We incorporate calibrated module-level uncertainty estimates with models of human workload to guide both decisions.
    \item We systematically evaluate several module selectors (e.g. direct optimization and graph-based methods) and querying algorithms (e.g. query-until-\allowbreak confident and workload-aware variants) in controlled synthetic experiments, revealing tradeoffs in system and user variable robustness, efficiency, and user workload.
    \item We demonstrate the approach in a real-world robot-assisted bite acquisition task, achieving higher recovery success and lower user querying workload compared to baselines in two in-lab studies with emulated limitations and an in-home study involving users with real mobility limitations.
\end{itemize}
}

\vspace{-3ex}
\section{Related Work}

\indent\indent\textbf{Anomaly detection and recovery in robotics.} 
Robotic policies are prone to failure during execution, especially in unstructured environments or when encountering novel objects. Despite recent advances in end-to-end robot learning, these approaches can be difficult to interpret \cite{agia2025unpacking, xu2025can, duan2024aha}. In contrast, modular methods can offer more predictable and controllable performance in certain scenarios. Our work focuses on full-stack modular methods that consider failure detection and recovery across different modules of a robot system. While some human-in-the-loop approaches \cite{ahmad2024adaptable, fitzgerald2022inquire, larianlearner} assume that the system is in a failure state where recovery is required, our approach includes failure detection using confidence estimates.

It is natural to represent such problems using a graph representation, as in prior online approaches for robot failure detection and recovery \cite{ahmad2024adaptable, cornelio2024recover, pan2022failure}. The overall interaction process can also be modeled by certain types of graphs, ranging from finite-state machines \cite{cornelio2024recover}, MDPs/POMDPs \cite{garrabe2024enhancing, pan2022failure}, and Behavior Trees \cite{ahmad2024adaptable, colledanchise2018behavior, mayr2022skill}. Although graph representations have been widely studied in prior work, our method goes further by not only modeling systems using graphs but also leveraging graph algorithms to select which modules to query for intervention. Two of our four proposed methods for selecting which modules to query (which we denote module selectors) are graph-based.


\indent\textbf{Robots that ask for help.}
Robot systems may decide to query humans on the basis of several different criteria. Some approaches ask for help when a failure is detected after a robot action has been executed \cite{knepper2015recovering, tellex2014asking}, some based on a confidence estimate about potential robot actions to take \cite{ren2023robots, mullen2024towards}, and others based on an estimate about the potential information gain that could arise from soliciting human feedback \cite{papallas2022ask, fitzgerald2022inquire}. 
Departing from existing work, our query rules depend on both module confidence values and estimated human workload costs of querying. 

The types of feedback the robot can ask for can also vary, ranging from natural language \cite{cakmak2012designing, ren2023robots, shi2024yell}, to actions \cite{knepper2015recovering, tellex2014asking, vats2025optimal}, to labels \cite{cakmak2012designing}. 
Our module selection methods are agnostic to the specific types of feedback. Unlike prior work that queries for a single action-selection module \cite{ren2023robots, karli2025ask} or reward model \cite{fitzgerald2022inquire}, we consider querying for multiple modules in a full-stack policy architecture.

\vspace{-2ex}
\section{Problem Formulation}
\label{sec:problem-formulation}

\textbf{Module graph.} We assume that our robot policy architecture $\mathcal{P}: S \rightarrow A$ is structured as a module graph $G_M$, which describes the relationships between the $N$ modules in the architecture. Here, $S$ represents the policy input state (e.g. RGB-D images), and $A$ represents the policy action output (e.g. robot end-effector pose). The vertices in $G_M$ are modules, and an edge exists from one module $M_i$ to another module $M_j$ if the output of $M_i$ is an input to $M_j$.

A \emph{module} is a tuple $M = (\pi, \mathcal{X}, \mathcal{Y}, c, q(t))$ where $\pi: \mathcal{X} \to \mathcal{Y}$ is a policy function from input space $\mathcal{X}$ to output space $\mathcal{Y}$, $c:\mathcal X\to \mathbb [0,1]$ is a confidence score, and $q(t)$ is a time-dependent cost for querying the human about the module. For instance, a module could be a perception module that extracts geometric information about the robot's environment from RGB-D input, or a planning module that extracts a robot trajectory given obstacle states in the environment. We assume that one module in $G_M$ (which we denote $M_1$) takes in the state $s \in S$ (i.e. $\mathcal X= S$), and one module (which we denote $M_N$) produces a final action $a \in A$ (i.e. $\mathcal Y= A$), with other modules having arbitrary input/output spaces.

\textbf{Research problem.} Our goal is to decide what and when to query,
which we decompose into (1) a module selector algorithm $\psi_{ms}$ which chooses a single module, and (2) a querying algorithm $\psi_q$ that decides whether to query or to execute the autonomous action $a=\mathcal{P}(s) \in \mathcal{A}$. For a particular state $s \in S$, the module selector algorithm $\psi_{ms}$ and querying algorithm $\psi_q$ should minimize $J(\psi_{ms}, \psi_q)$, the weighted sum of the query cost $J_{\text{query}}(t;\psi_{ms}, \psi_q)$ and the task cost $J_{\text{task}}(t;\psi_{ms},\psi_{q})$, over the interaction horizon $T$:

\vspace{-3.5ex}

\begin{align} \label{eq:objective}
    J(\psi_{ms}, \psi_q) &= \frac{1}{T} \sum_{t=1}^T \left( w J_{\text{query}} + (1-w) J_{\text{task}} \right) \nonumber \\
    &= \frac{1}{T} \sum_{t=1}^T \left( w \sum_{M_i \in \Omega(t)} q_i(t) + (1-w) (1-r(t)) \right) 
\end{align}

where $w \in [0,1]$ determines the trade-off between minimizing human workload and recovering from failures efficiently, $\Omega(t)$ denotes the set of modules for which we query at time $t$, $M_i$ refers to an individual module in this set, $q_i(t)$ is the query cost for module $i$, and $r(t)$ is the binary task reward, which is equal to 1 if the execution succeeds, and 0 otherwise.

\textbf{Human feedback.} We assume that for every module $M_i$, the user can provide feedback $f \in \mathcal{Y}_i$ by choosing a possible output of that module (e.g. classifying an object or selecting a high level skill). When we receive this feedback, we replace the module output with the expert feedback $f$, along with an expert confidence $c=c_{\text{expert}}$. 

\textbf{Module redundancy.} The overall task reward $r(t)$ depends on the success of individual modules. The modules can be structured in two ways, depending on the robot system architecture:

\begin{itemize}[topsep=0pt, leftmargin=*]
    \item A \textit{redundant} manner, 
    where at least one module $M_i$ must succeed for their combination to succeed, so success is a union. For example, a policy architecture may include two redundant modules that predict object material properties: an off-the-shelf foundation model, and a domain-specific neural network \cite{wu2025savor}.
    \item A \textit{non-redundant} manner, 
    where all modules $M_i$ must succeed for their combination to succeed, so success is an intersection. For example, a goal-reaching manipulation policy architecture may require two modules to succeed: a goal inference module, and a planning module that infers a trajectory to the goal \cite{ma2024hierarchical}.
\end{itemize}
The overall success $r(t)$ depends on redundant and non-redundant combinations of modules, based on the robot system architecture.

\vspace{-2ex}
\section{Module Selectors}
\label{sec:module-selector}

The module selector algorithms $\psi_{ms}$ select which module to query by considering the 1-timestep version of the overall objective (Eq. \ref{eq:objective}). While we cannot determine \emph{a priori} whether each module $M_i$ is in failure or success state, we have access to confidences $c_i$. We consider three \textit{proxy objectives} to minimize:

\textit{Proxy objective 1} (product-of-confidences):
\vspace{-1ex}
\begin{equation} \label{eq:proxy-1}
    \textstyle w \sum_{M_i \in \Omega(t)} q_i(t) + (1-w) (1-\prod_{M_i \notin \Omega(t)} c_i)
\end{equation}
\vspace{-1ex}
\textit{Proxy objective 2} (sum-of-uncertainties):
\begin{equation} \label{eq:proxy-2}
    \textstyle w \sum_{M_i \in \Omega(t)} q_i(t) + (1-w) \sum_{M_i \notin \Omega(t)} (1-c_i)
\end{equation}
\vspace{-1ex}
\textit{Proxy objective 3} (redundancy-dependent):
\begin{align} \label{eq:proxy-3}
    &\textstyle w \sum_{M_i \in \Omega(t)} q_i(t) + (1-w) (1 - \hat r(\Omega(t))
\end{align}

where reward estimate $\hat r(\Omega(t))$ combines sums (redundant) and products (non-redundant) of confidences for modules $M_i\notin \Omega(t)$ (see App.  \appref{sec:proxy-objectives} for justifications for how each proxy objective approximates $J_{\text{task}}$).

We consider four module selector algorithms---two direct proxy optimization algorithms (\textbf{Mixed-Integer Programming} and \textbf{Brute-Force}) and 
two graph-based algorithms (\textbf{Binary Tree Query} and \textbf{Graph Query}). We also consider three baseline module selectors, two of which do not consider query costs or confidence scores, and one that only considers confidence scores.
The baselines are (1) \textbf{Never Query} which does not choose any module to query, (2) \textbf{Topo Query} which simply selects the first module in the topological ordering of the module graph $G_M$ which has not been previously selected, and (3) \textbf{Confidence Query} which selects the module with the lowest confidence score.
\vspace{-2ex}
\subsection{Proxy Objective Optimization}

We consider two module selectors that directly use the redundancy-dependent proxy objective (Eq. \ref{eq:proxy-3}). The \textbf{Mixed-Integer Programming (MIP)} module selector directly minimizes this objective using a mixed-integer solver (SCIP \cite{bolusani2024scip}). The \textbf{Brute-Force} module selector evaluates this objective for querying each module $M_i$ in the module graph $G_M$, and then selects the module $M_i$ that achieves the minimal objective.

\begin{figure*}[!ht]
    {
    \tolerance=10
    \emergencystretch=2em
    \centering
    \includegraphics[width=0.9\linewidth,alt={Three diagrams, two which illustrate different module selection strategies, and one which shows four querying algorithms. The left diagram shows a binary tree querying module selector, where left edges represent querying, and right edges represent autonomous execution, annotated with query costs and confidence-based rewards. The center diagram depicts a graph query module selection strategy, showing a state-transition graph for execute-or-query decisions across success and failure states. The right diagram compare four querying algorithms: Execute-First, Query-then-Execute, Query-until-Confident, and Query-until-Confident-Workload-Aware.}]{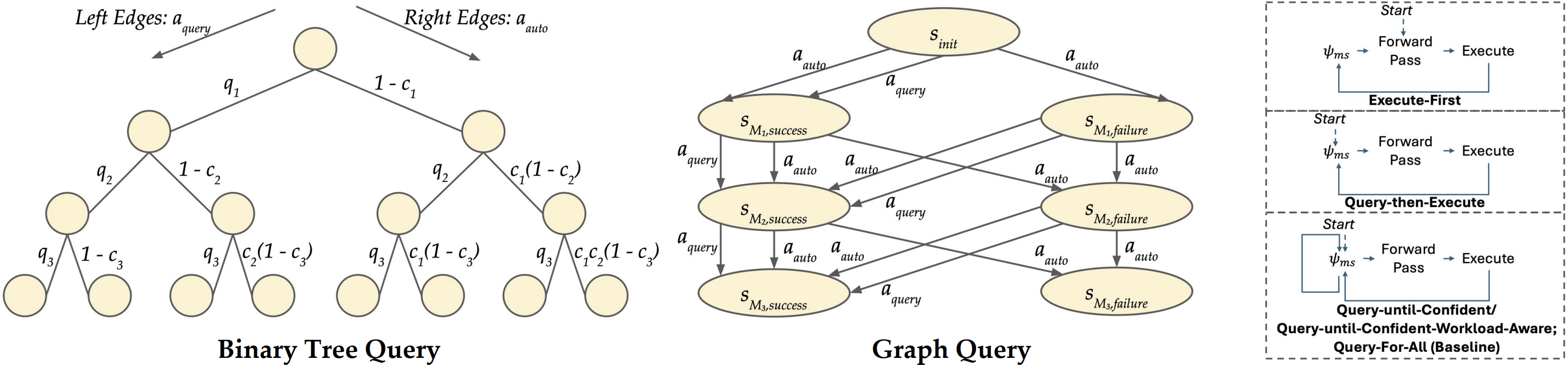}
    \vspace{-0.2cm}
    \caption{(left) \textsc{BinaryTreeQuery} graph example for $N{=}3$, (middle) \textsc{GraphQuery} graph example for $N{=}3$, (right) Querying algorithms and \textsc{Query-For-All} baseline, which decide when to query (calling module selector $\psi_{ms}$) and when to execute actions (calling \textsc{ForwardPass} to get action $a$, then calling \textsc{Execute}). Querying algorithms include \textsc{Execute-First}, which executes once prior to starting to query, \textsc{Query-then-Execute}, which alternates between querying and execution, and \textsc{Query-until-Confident}/\textsc{Query-until-Confident-Workload-Aware}, both of which repeatedly query until a stopping condition is met. The \textsc{Query-For-All} baseline queries for all modules before execution.}
    \Description{Three diagrams, two which illustrate different module selection strategies, and one which shows four querying algorithms. The left diagram shows a binary tree querying module selector, where left edges represent querying, and right edges represent autonomous execution, annotated with query costs and confidence-based rewards. The center diagram depicts a graph query module selection strategy, showing a state-transition graph for execute-or-query decisions across success and failure states. The right diagram compare four querying algorithms: Execute-First, Query-then-Execute, Query-until-Confident, and Query-until-Confident-Workload-Aware.}
    \label{fig:graph-and-schematics}
    }
\end{figure*}

\vspace{-2ex}
\subsection{Binary Tree Query}

This algorithm uses a binary tree graph to model the decision of querying each module. The edge weights are defined so that paths from the root to the leaves correspond to the product-of-confidences proxy objective (Eq.  \ref{eq:proxy-1}). We then treat the module optimization problem as a shortest-cost path problem, running Dijkstra's algorithm on the graph to obtain the module $M_i$.

The binary tree graph $B = (V,E)$ (shown in Fig. \ref{fig:graph-and-schematics} (left)) has edges $E$ corresponding to the actions of querying $a_{\text{query}}$ or acting autonomously $a_{\text{auto}}$ for module $M_i$, and vertices $V$ that encode the querying decisions of modules prior to module $M_i$ in the topological ordering of the module graph $G_M$. The edge costs for query actions are $C(e) = q_i(t)$, representing the query cost for module $M_i$. The edge costs for autonomous actions are chosen to create a telescoping sum with the correct product of confidences. In particular, $C(e) = (1-c_i) \prod_{j\in prev} c_j$, where $prev$ are prior autonomous modules on the path from root to $M_i$, and if $prev$ is empty, $C(e) = 1 - c_i$.

\vspace{-2ex}
\subsection{Graph Query}

This algorithm constructs a directed graph representing the modular policy and treats module selection as a shortest path problem, running Dijkstra's algorithm on the graph to select the module $M_i$ (equivalent to optimizing the sum-of-uncertainties objective, Eq. \ref{eq:proxy-2}).

Given the topological ordering of modules in the module graph $G_M$, the graph $G = (V,E)$ is shown in Fig.  \ref{fig:graph-and-schematics} (middle), where vertices $V$ correspond to the state of each module, and edges $E$ correspond to actions, both query $a_{\text{query}}$ and autonomous $a_{\text{auto}}$. The vertices $V$ consist of an initial state $s_{\text{init}}$ and pairs of success and failure states for each module $M_i$ ($s_{M_i,\text{success}}$ and $s_{M_i,\text{failure}}$). 
The edge costs for query actions are a scaled version of the query cost $C(e) = \epsilon q_i(t)$, with scaling factor $\epsilon>0$ (for module selectors). The edge costs for autonomous actions are the module uncertainties $C(e) = 1-c_i$.

\vspace{-2ex}
\section{Querying Algorithms}
\label{sec:querying-algorithms}

Querying algorithms decide whether to make a query or execute the action $a$ produced by the final action module $M_N$ (Sec. \ref{sec:problem-formulation}). We consider four querying algorithms and one baseline, illustrated in Fig. \ref{fig:graph-and-schematics} (right).
They use the following primitives:

\begin{itemize}[topsep=0pt, leftmargin=*]
    \item \textsc{ForwardPass}: represents passing the current state $s \in \mathcal{S}$ and query set $\Omega(t)$ through all of the modules in the module graph $G_M$ to yield the final action $a$ and updated confidences. In the manipulation setting, this could represent passing the current perceptual state (e.g. RGB-D camera data) to the policy architecture, producing the desired end-effector pose, along with  confidences for all modules in the policy architecture.
    \item \textsc{Execute}: represents executing the action $a$ and observing overall boolean $outcome$. In the manipulation setting, this could represent commanding the robot to the desired end-effector pose and assessing whether the manipulation task was successful.
\end{itemize}

All querying algorithms call the module selector $\psi_{ms}$ to select a candidate module to query, but differ in how many queries are made prior to execution. The \textbf{Execute-First} algorithm initially executes the autonomous action $a$ from the \textsc{ForwardPass} primitive. If $a$ fails, it alternates between calling the module selector $\psi_{ms}$ and executing until success.
\textbf{Query-then-Execute} always alternates between calling the module selector $\psi_{ms}$ and executing until success.

\begin{algorithm}
\caption{\textsc{Execute-First} and \textsc{Query-then-Execute}.}
\small
\begin{algorithmic}[1]
\If{\textsc{Execute-First}}
\State $a = \textsc{ForwardPass}(\emptyset)$
\State $outcome=$ \textsc{Execute}$(a)$
\ElsIf{\textsc{Query-then-Execute}}
\State $outcome= $\textsc{ FALSE}
\EndIf
\While{$outcome= $\textsc{ FALSE}}
    \State $\Omega(t) = \psi_{ms}(G_M)$
    \State $a=$ \textsc{ForwardPass}$(\Omega(t))$
    \State $outcome=$ \textsc{Execute}$(a)$
\EndWhile
\end{algorithmic}
\end{algorithm}

\begin{algorithm}
\caption{\textsc{Query-until-Confident} and \textsc{Query-until\\-Confident-Workload-Aware}; \textsc{Query-For-All} baseline.}
\small
\begin{algorithmic}[1]
\State $outcome= $\textsc{ FALSE}
\While{$outcome= $\textsc{ FALSE}}
    \Repeat
        \State $\Omega(t) = \psi_{ms}(G_M)$
        \If{\textsc{Query-until-Confident}}
        \State $StopCondition=[\hat{r}(\Omega(t)) > \tau]$
        \ElsIf{\textsc{Query-until-Confident-Workload-Aware}}
        \State $StopCondition=[c_{\text{expert}} - c_{i} < \lambda q_i(t)]$
        \ElsIf{\textsc{Query-For-All}}
        \State $StopCondition=[\Omega(t) = \emptyset]$
        \EndIf
    \Until{$StopCondition= $\textsc{ TRUE}}
    \State $a=$ \textsc{ForwardPass}$(\Omega(t))$
    \State $outcome=$ \textsc{Execute}$(a)$
\EndWhile
\end{algorithmic}
\end{algorithm}

\textbf{Query-until-Confident}
repeatedly calls $\psi_{ms}$ until the proxy task reward estimate $\hat r(\Omega(t))$ (Eq. \ref{eq:proxy-3}) exceeds a certain threshold $\tau$, then executes the action $a$.
\textbf{Query-until-Confident-Workload-Aware}
repeatedly calls $\psi_{ms}$ until the confidence gain due to querying module $M_i$, $c_{\text{expert}} - c_{i}$ ($c_{\text{expert}}$ defined in Sec. \ref{sec:problem-formulation}), is less than the scaled cost of querying $\lambda q_i(t)$, for scaling factor $\lambda>0$ (for querying algorithms). Finally, the \textbf{Query-For-All}
baseline repeatedly calls $\psi_{ms}$ until it has queried for all modules, then executes the resultant action $a$.

\vspace{-2ex}
\section{Synthetic Simulation: Systematic Ablations}
\label{sec:synthetic-results}

We develop a synthetic module simulation to investigate how system and user variables affect module selector and querying algorithm performance (Secs. \ref{sec:module-selector}, \ref{sec:querying-algorithms}). Our simulation uses $N$ modules connected via a random module graph $G_M$, where modules are implemented as logic gates (AND/OR) operating on Boolean inputs (including a Boolean state $s$). Individual modules are set to be in either a success or failure state: modules in success states output their logic gate value, while modules in failure states always output \textsc{False}. See App.  \appref{app:synthetic-exp-setup} for hyperparameter details (including proxy objective weighting and \textsc{GraphQuery} $\epsilon$).

\textbf{Independent variables.} We examine four variables covering key system and user factors that affect robot and human-robot interaction performance (full settings in App.  \appref{app:synthetic-exp-setup}). We consider three system variables:
\vspace{-0.5ex}
\begin{itemize}[leftmargin=*]
    \item \textbf{The number of modules $N$ (Sec. \ref{subsec:num-modules}).} The number of modules in a robot policy architecture can vary, ranging from simple modular perception, planning, and control architectures, to more complex modular architectures involving up to 100 modules.
    \item \textbf{Module redundancy/graph structures $G_M$ (Sec. \ref{subsec:graph-structure}).} Robot policy architectures include both redundant and non-redundant structures (Sec. \ref{sec:problem-formulation}). Different module selectors have different redundancy assumptions based on their proxy objective. We consider 4 redundancy structures with varied module types and orderings: fully redundant (all-OR), fully dependent (all-AND), fully redundant followed by fully dependent (OR-then-AND), fully dependent followed by fully redundant (AND-then-OR).
    \item \textbf{Confidence values $c_i$ (Sec. \ref{subsec:confidence}).} Robot policy architectures can have a diversity of confidence scores across the modules. We assign either a high or a low confidence value to each module in $G_M$. We additionally treat a module's confidence value as the probability that the module is in a success state.
\end{itemize}

And we consider one user variable:    

\begin{itemize}[leftmargin=*]
    \item \textbf{Query costs $q_i$ (Sec. \ref{subsec:query-cost}).} The costs of querying for each module in a policy architecture can vary, due to the diversity of feedback types. We consider different values of uniform query costs $q_i$ for all modules.
\end{itemize}    

We vary each independent variable in isolation. When a variable is not varied, we fix its value as follows: (1) $N{=}10$ modules, (2) Fully dependent (all-AND) module structure, (3) 3 modules with confidence $0.1$ and all other modules with confidence $1$, (4) Uniform query cost of $0.32$. We also assume no noise in the expert ($c_{\text{expert}}{=}1.0$). Additionally, we use a fixed querying algorithm (\textsc{Query-Until-Confident-Workload-Aware}) when analyzing module selectors, and a fixed module selector (\textsc{GraphQuery}) when analyzing querying algorithms.

\noindent \textbf{Metrics}. We evaluate the methods using 5 metrics (lower is better):
\begin{itemize}[leftmargin=*]
    \item \textbf{Task Cost.} 0 if the agent successfully recovered from the failure after $T$ timesteps, 1 otherwise. This represents whether the robot could recover from the failure given its time horizon constraint.
    \item \textbf{Query Cost.} $\sum_{t=1}^T b_t q_i(t)$, where $T$ is the number of timesteps needed to achieve success or the maximum time horizon (which is proportional to the number of modules $N$), and $b_t$ is a binary indicator variable for whether the robot queried at timestep $t$. This represents the total user workload used to recover from the robot's failure.
    \item \textbf{Number of Failed Attempts.} $\sum_{t=1}^T g_t$, where $g_t$ is a binary indicator variable for whether a failed execution occurred at timestep $t$. This represents the number of failed robot executions encountered before the robot recovered from the failure.
    \item \textbf{Computation Time.} $\sum_{t=1}^T (Time(t))$, where $Time(t)$ is the total runtime between successive \textsc{EXECUTE} statements. This represents the total computation time used by the robot to run the module selector and the querying algorithm at each timestep (not including the time required to query).
    \item \textbf{Total Timesteps $T$.} This includes both the number of queries and the number of failed executions encountered before the robot recovered from the failure.
\end{itemize}

\begin{figure*}[!ht]
    \centering
    \includegraphics[width=0.85\linewidth,alt={On the left side, multi-panel plots comparing module selectors and querying algorithms in a synthetic modular setting. Across varying redundancy, confidence levels, workloads, and number of modules, confidence-based and workload-aware strategies achieve lower task cost outcomes. Effects differ between low- and high-variance regimes. On the right-side, multi-panel plots comparing graph-based and confidence-based module selectors, showing that the graph-based module selectors have a lower task cost metric across varying confidence levels.}]{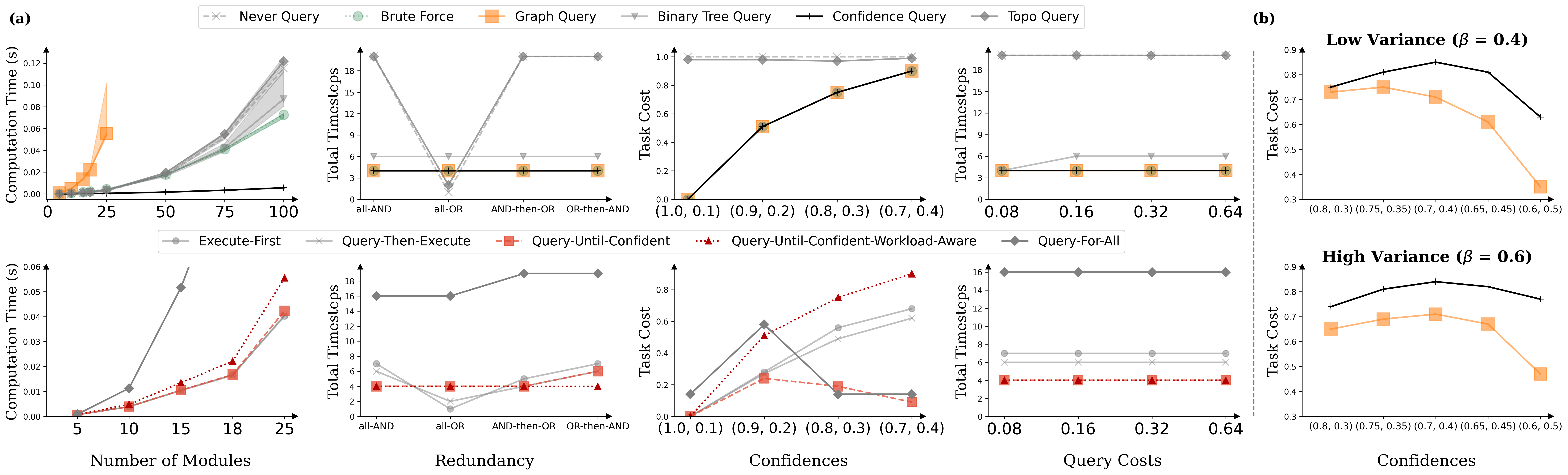}
       \vspace{-0.25cm}
    \caption{(a) Systematic ablation experiments, showing the 4 simulation independent variables: (1) number of modules, (2) graph redundancies, (3) confidences, (4) query costs, with median values across 100 trials (mean for Task Cost)\protect\footnotemark. We find that the \textsc{BruteForce} and \textsc{GraphQuery} module selectors (and \textsc{ConfidenceQuery} baseline) are the most robust to varying redundancy, confidences, and query costs, with \textsc{BruteForce} and \textsc{ConfidenceQuery} being the most scalable. Additionally, we find that the \textsc{Query-until-Confident} and \textsc{Query-until-Confident-Workload-Aware} querying algorithms are the most robust across redundancy and query costs, with \textsc{Query-until-Confident} having the best scalability and robustness to confidences; (b) Module heterogeneity experiments. We find that \textsc{GraphQuery} outperforms \textsc{ConfidenceQuery}, particularly when module confidences overlap and workload variance $\beta$ is high. Detailed results in App.  \appref{app:synthetic-exps}.}
    \Description{On the left side, multi-panel plots comparing module selectors and querying algorithms in a synthetic modular setting. Across varying redundancy, confidence levels, workloads, and number of modules, confidence-based and workload-aware strategies achieve lower task cost outcomes. Effects differ between low- and high-variance regimes. On the right-side, multi-panel plots comparing graph-based and confidence-based module selectors, showing that the graph-based module selectors have a lower task cost metric across varying confidence levels.}
    \label{fig:synthetic-full}
\end{figure*}

Fig. \ref{fig:synthetic-full}(a) compares the algorithm performance for each variable, highlighting one metric\fnref{note-exclusions} (full results detailed in App.  \appref{app:synthetic-exps}). We additionally highlight results in App.  \appref{appendix:noisy-expert} for varying an additional user variable, the expert confidence level $c_{\text{expert}}$ (Sec. \ref{sec:problem-formulation}), where we find that the \textsc{BruteForce} and \textsc{GraphQuery} module selectors and both \textsc{Query-Until-Confident} querying algorithms perform best in high expert confidence regimes, with performance degrading as the confidence decreases.

\footnotetext{\label{note-exclusions}Because the computation time of the \textsc{GraphQuery} exceeded 1 second for graph sizes $N \geq 50$, we omit the computation time for this module selector. Additionally, we omit reporting MIP results for all graph sizes because its computation time also exceeds this threshold.}

\vspace{-2ex}
\subsection{Varying number of modules $N$}
\label{subsec:num-modules}

\textbf{Module selectors.} All metrics except \textit{Computation Time} are invariant to $N$, for all methods except \textsc{NeverQuery} and \textsc{TopoQuery}. \textsc{ConfidenceQuery} is the most scalable method with linear time complexity in $N$ (Fig. \ref{fig:synthetic-full}(a)), making it most computationally efficient for large policy architectures. As \textit{Computation Time} scales with $T$, methods requiring many timesteps (\textsc{NeverQuery} and \textsc{TopoQuery}) incur higher computation time.

\textbf{Querying algorithms.} All metrics except \textit{Computation Time} are invariant to $N$, for all methods except \textsc{Query-For-All}, with \textit{Task Cost} and \textit{Query Cost} of 0 and 0.96, respectively. We note that \textsc{Query-Until-Confident-Workload-Aware} has the highest \textit{Computation Time} (Fig. \ref{fig:synthetic-full}(a)), due to the additional step to predict the querying cost, while the other three methods have nearly-identical values for \textit{Computation Time}. For large robot policy architectures, we recommend not selecting \textsc{Query-Until-Confident-Workload-Aware} if runtime is crucial.

\vspace{-2ex}

\subsection{Varying graph structure $G_M$}
\label{subsec:graph-structure}

\textbf{Module selectors.} All metrics are lower for the all-OR structure across methods (except \textit{Query Cost}, which is invariant to the structure). For this redundant structure, there is little benefit to ask for help to recover from failures, leading \textsc{NeverQuery} and \textsc{TopoQuery} to perform well in \textit{Total Failed Attempts} and \textit{Total Timesteps}. Method rankings (Fig. \ref{fig:synthetic-full}(a)) remain invariant for the other redundancy structures. \textsc{GraphQuery} is competitive with the other methods, even when its proxy objective does not match the redundancy structure.

\textbf{Querying algorithms.} All metrics are lower for the all-OR structure across methods. In general, both \textsc{Query-until-Confident} variants have the lowest \textit{Total Failed Attempts}, except in the all-OR setting (where \textsc{Execute-First} and \textsc{Query-then-Execute} outperform them). Thus, if the robot policy architecture is not fully redundant, we recommend either \textsc{Query-until-Confident} variant to maximize recovery efficiency. 

\vspace{-2ex}
\subsection{Varying confidence values $c_i$}
\label{subsec:confidence}

\textbf{Module selectors.} All metrics (e.g. \textit{Task Cost}, Fig. \ref{fig:synthetic-full}(a)) are higher across methods when confidences overlap. As \textsc{BruteForce}, \textsc{GraphQuery}, and \textsc{ConfidenceQuery} are the most competitive methods regardless of confidence level, we recommend any of them.

\textbf{Querying algorithms.} All metrics (e.g. \textit{Task Cost}, Fig. \ref{fig:synthetic-full}(a)) are higher across methods when confidences overlap, except for \textsc{Query-Until-Confident} and \textsc{Query-For-All}. We find that the \textsc{Query-Until-Confident} method is the most robust to confidence variations. We recommend any method if the confidence scores are well-separated in the robot policy architecture, and \textsc{Query-Until-Confident} if confidence values are overlapping.

\vspace{-2ex}
\subsection{Varying query costs $q_i$}
\label{subsec:query-cost}

\textbf{Module selectors.} All metrics except \textit{Task Cost} are invariant to $q_i$, with \textsc{BruteForce}, \textsc{GraphQuery}, and \textsc{ConfidenceQuery} having lower \textit{Total Timesteps} (Fig. \ref{fig:synthetic-full}(a)) compared to the other module selectors. Regardless of the level of user querying workload, we recommend either of these module selectors. 

\textbf{Querying algorithms.} Besides \textit{Query Cost}, all metrics are largely invariant to $q_i$ across querying algorithms. As both \textsc{Query-Until-Confident} variants have the lowest \textit{Total Failed Attempts} and generally lower \textit{Total Timesteps} (Fig. \ref{fig:synthetic-full}(a)), we recommend either of these variants.

\vspace{-2ex}
\section{Synthetic Simulation: Module Heterogeneity}
\label{sec:synthetic-results-hetero}

{
\tolerance=2000
\emergencystretch=2em
We now consider a simulation setting with an additional user variable, \textbf{query cost variance $\beta$}, allowing query costs $q_i$ to vary across modules. This models more realistic cost variation across feedback types. We sample each $q_i$ uniformly around a nominal value $Q$, i.e., $q_i{ \sim }\text{Uniform}[(1{-}\beta)Q, (1{+}\beta)Q]$, with $Q{=}0.32$ (the query cost value used when not varied in Sec. \ref{sec:synthetic-results}). We compare the best performing module selectors from Sec. \ref{sec:synthetic-results} based on \textit{Task Cost}: the query cost-aware method \textsc{GraphQuery}, and the confidence-only baseline \textsc{ConfidenceQuery} (both with \textsc{Query-Until-Confident-Workload-Aware}). 
}

We find that \textsc{GraphQuery} consistently outperforms \textsc{ConfidenceQuery} in the \textit{Total Timesteps} and \textit{Task Cost} metrics, especially when module confidences overlap or workload variance is high (\textit{Task Cost} in Fig. \ref{fig:synthetic-full}(b)). Under these conditions, upstream modules with lower confidence and high workload may succeed, whereas downstream modules with higher confidence but low workload may fail. \textsc{ConfidenceQuery} incorrectly selects these upstream modules first, whereas \textsc{GraphQuery} correctly selects the downstream module (full results in App.  \appref{appendix:cq-gq-comprison}).

\vspace{-0.5\baselineskip}
\section{Robot-Assisted Bite Acquisition Experiments}

Our real robot experiments use a robot-assisted bite acquisition architecture with $N = 4$ modules: food type identification (GPT-4o), bounding box selection (GroundingDINO), skill selection (GPT-4o), and skill parameter selection (RT-1). The first three modules are VLMs, while the skill selection module is a VLA. Food type identification and skill selection support learning from feedback via retrieval-augmented generation (RAG) (architecture and RAG details in App.  \appref{sec:bite-acq-baseline}). We develop calibrated confidence scores for each module using a population-based interval procedure (detailed in App.  \appref{sec:calibration-data-confidence-intervals}-\ref{subsec:calibrated-confidence-scores}). We estimate module query costs using a predictive workload model from prior work \cite{banerjee2024ask} (App.  \appref{app:workload-model}).

To select a module selector and querying algorithm for bite acquisition, we adopt the recommendation from the closest synthetic setting: we use $N{=}10$ to approximate system scale; we model bite acquisition as non-redundant (all-AND) as all modules must be correct for success (Sec. \ref{sec:problem-formulation}); we use the confidence setting $(1, 0.1)$ to match our binary calibrated module confidence scores (App.  \appref{subsec:calibrated-confidence-scores}); we select $q_i{=} 0.32$ to match the empirical mean of workload model predictions; we assume $c_{\text{expert}}{=}1.0$ as expert feedback is available for all modules. Under these conditions (Sec. \ref{sec:synthetic-results}) and the heterogeneity experiments (Sec. \ref{sec:synthetic-results-hetero}), we use \textsc{GraphQuery} and \textsc{Query-Until-Confident-Workload-Aware}, which minimized \textit{Total Timesteps} and \textit{Task Cost}.\footnote{While \textsc{GraphQuery} and \textsc{BruteForce} were equally competitive, \textsc{BruteForce} produced false-positive queries with our binary confidence scores. We chose \textsc{Query-until-Confident-Workload-Aware} over \textsc{Query-until-Confident} for better generalization to time-varying workload.}

\begin{figure*}[!th]
    \centering
    \includegraphics[width=0.7\linewidth,alt={Composite figure showing the experimental setup and study contexts for robot-assisted feeding. The left panel depicts the hardware configuration, including a Kinova Gen3 robot with a fork end-effector, depth camera, and tablet-based user interface. The center panels show example meal plates and participants in in-home studies with mobility limitations interacting with the system. The right panel presents a grid of in-lab study participants with emulated mobility limitations seated at a table while the robot performs feeding tasks.}]{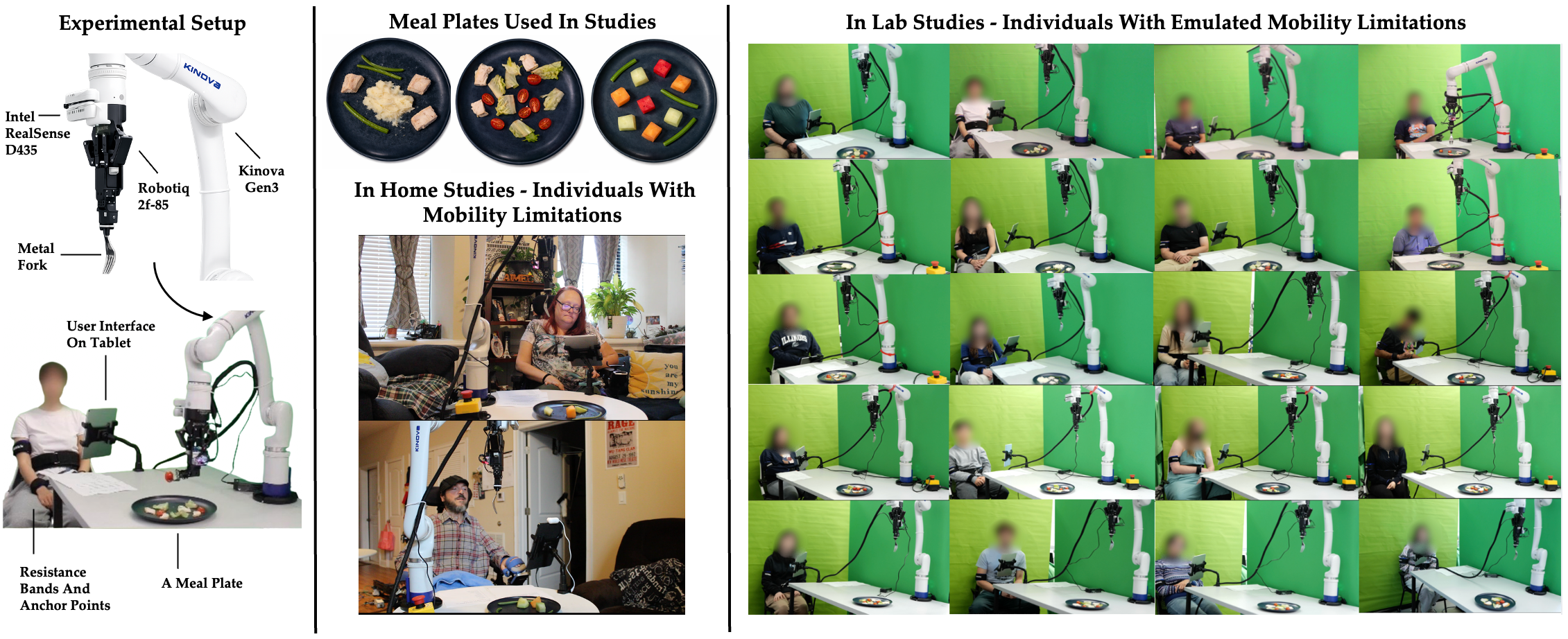}
    \vspace{-0.25cm}
    \caption{Experimental setup. (left) Robot and study setup; (middle, top) Meal plates (``Thanksgiving meal", ``savory salad", ``mixed salad"); (middle, bottom) In-home users with mobility limitations; (right) In-lab users with emulated mobility limitations.}
    \Description{Composite figure showing the experimental setup and study contexts for robot-assisted feeding. The left panel depicts the hardware configuration, including a Kinova Gen3 robot with a fork end-effector, depth camera, and tablet-based user interface. The center panels show example meal plates and participants in in-home studies with mobility limitations interacting with the system. The right panel presents a grid of in-lab study participants with emulated mobility limitations seated at a table while the robot performs feeding tasks.}
    \label{fig:experimental-design}
\end{figure*}

\vspace{-0.25ex}

\begin{figure*}[!th]
    \centering
    \includegraphics[width=0.8\linewidth,alt={Multi-panel bar charts showing in-lab and in-home real-robot user study results. Across subjective and objective metrics, the proposed method reduces user workload and improves task success compared to always-query and never-query baselines, with statistically significant differences indicated for several metrics.}]{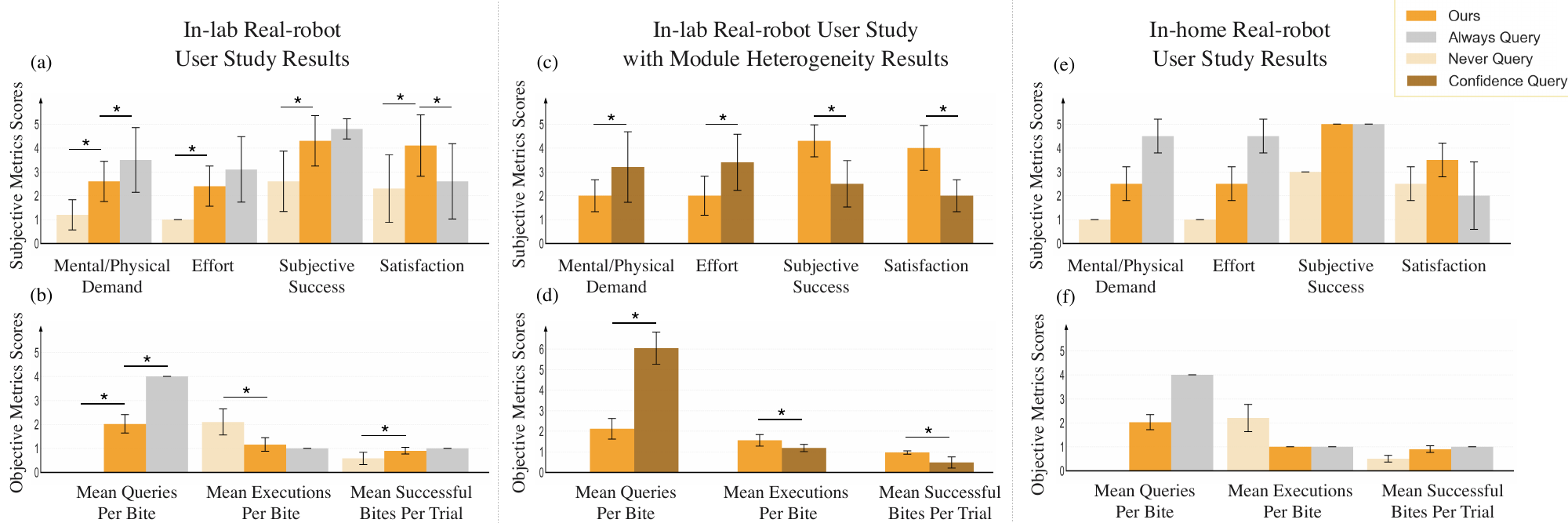}
    \vspace{-0.25cm}
    \caption{User study metrics. In-lab real-robot study: (a) subjective scores, (b) objective scores; In-lab real-robot study with module heterogeneity: (c) subjective scores, (d) objective scores; In-home real-robot study: (e) subjective scores, (f) objective scores ($*$ indicates statistical significance p < 0.05.).}
    \Description{Multi-panel bar charts showing in-lab and in-home real-robot user study results. Across subjective and objective metrics, the proposed method reduces user workload and improves task success compared to always-query and never-query baselines, with statistically significant differences indicated for several metrics.}
    \label{fig:user-study-figure}
\end{figure*}

\textbf{User Studies.} We conducted three IRB-approved studies to evaluate algorithm task success and querying workload: two in-lab studies, and one in-home study involving two individuals with severe mobility limitations. All studies use a Kinova Gen3 6-DoF arm with a Robotiq 2F-85 gripper, and we replicate a custom-designed feeding tool for the studies \cite{jenamani2024flair}. A key methodological feature of both of in-lab studies, supporting ecological validity, is the emulation of mobility constraints in participants without pre-existing mobility impairments using occupational therapy resistance bands \cite{liu2025grace} (Fig. \ref{fig:experimental-design} (left)). 

For each method, the robot acquires 5 food items from a plate, with a maximum of 3 acquisition attempts per item for a reasonable study duration. The robot may ask 4 types of queries, one for each bite acquisition module (App. \appref{sec:bite-acq-queries}). The method and plate sequences are counterbalanced across users in each study (Fig. \ref{fig:experimental-design} (middle, top)).

\textbf{Metrics}.
We report 4 subjective metrics (Mental/Physical Demand, Effort, Subjective Success, Satisfaction) and 3 objective metrics (Mean Queries/Executions/Successful Bites Per Plate), detailed in App.  \appref{sec:user-study-metrics}.

\vspace{-0.5\baselineskip}
\subsection{In-lab real-robot user study}
\label{subsec:in-lab-1}
{
\fussy
We first conducted an in-lab study with 10 participants without mobility limitations (7 male, 3 female; ages 19–33; 40\% with prior robot experience) (Fig. \ref{fig:experimental-design} (right)). We compared our method (\textsc{GraphQuery} and \textsc{Query-Until-Confident-Workload-Aware}) against two additional baselines: \textit{Never Query} (\textsc{NeverQuery} and \textsc{Query-\allowbreak then-\allowbreak Execute}) and \textit{Always Query} (\textsc{TopoQuery} and \textsc{Query-For-All}). The study evaluated whether our method improved task success over \textit{Never Query} and reduced querying workload over \textit{Always Query}. Participants interacted with two plates: a ``savory salad" (chicken, lettuce, cherry tomatoes) and a ``Thanksgiving meal" (chicken, green beans, mashed potatoes). Our method achieved the highest user satisfaction, significantly lower mental/physical workload than \textit{Always Query}, and higher subjective success than \textit{Never Query} (Wilcoxon signed-rank test, $\alpha{=}0.05$; Fig. \ref{fig:user-study-figure}(a–b)). Our method is more efficient than \textit{Always Query}, with higher task success than \textit{Never Query}.
}

\vspace{-1\baselineskip}
\subsection{In-lab real-robot user study with module heterogeneity}
\label{subsec:in-lab-2}

To distinguish our method from \textsc{ConfidenceQuery}, we introduced module heterogeneity into our setting by incorporating a faulty food detector $M_1$ (App.  \appref{sec:bite-acq-hetero}), similar to our second simulation study (Sec. \ref{sec:synthetic-results-hetero}). We conducted a second in-lab study with 10 additional participants (3 male, 6 female, 1 non-binary; ages 20-30; 30\% with prior robot experience) using the ``savory salad" plate (Fig. \ref{fig:experimental-design} (right)). We evaluated whether our method improved task success and querying workload compared to the \textit{Confidence Query} baseline (\textsc{ConfidenceQuery} module selector and \textsc{Query-Until-Confident-Workload-Aware} querying algorithm). Our method achieved significantly higher task performance, lower workload, and higher satisfaction than \textit{Confidence Query} (Wilcoxon test, $\alpha{=}0.05$; Fig. 5(c-d)), reinforcing the value of principled uncertainty and workload integration.

\vspace{-0.75\baselineskip}
\subsection{In-home real-robot user study}

We conducted an in-home user study on 2 individuals with mobility limitations requiring feeding assistance (Fig. \ref{fig:experimental-design} (middle, bottom)): one female, 48 years old, who has had Multiple Sclerosis for 17 years, and one male, 47 years old, who has been paralyzed with a C4-C6 spinal cord injury for 27 years. Based on the in-lab results (Secs. \ref{subsec:in-lab-1}, \ref{subsec:in-lab-2}), we compared our method to \textit{NeverQuery} and \textit{AlwaysQuery}. We evaluated the methods on a ``mixed salad" plate with additional food item diversity (watermelons, cantaloupes, honeydew, green beans). We find that the mental/physical demand and effort of our method are lower compared to \textit{Always Query}, and that users rated our method as more successful compared to \textit{Never Query} (Fig. \ref{fig:user-study-figure}(e-f)). In general, our method achieves the best user satisfaction score. 

Both users expressed that they liked the system, with higher satisfaction scores for our method than for both baselines. One user's absolute satisfaction level was reduced by the need for physical interaction with the interface and noted that a more accessible option, such as voice control, would significantly improve their experience. Because they regularly use a tablet interface in everyday interactions, answering a few more queries on the tablet did not require significantly extra effort for the study duration. However, they still rated \textit{AlwaysQuery} as having higher effort than our method. The other user mentioned that longer robot execution times (particularly when failures occurred) were a source of frustration that increased overall workload. 

\vspace{-1\baselineskip}
\section{Discussion}

Our framework readily extends to other modular robot systems, which we illustrate by considering two additional domains: a feeding architecture with visuo-haptic perceptual redundancy \cite{wu2025savor}, and a large multi-robot swarm system \cite{rubenstein2014programmable}. The feeding system maps naturally to an OR-then-AND redundancy structure (due to the perceptual redundancy) with low query cost ($q_i = 0.16$), yielding the same recommendation as our primary setting (Fig. \ref{fig:synthetic-full}). We can model the swarm domain with $N=100$ to capture its scale and an all-OR redundancy structure, leading to the \textsc{ConfidenceQuery} module selector for its superior scalability (Fig. \ref{fig:synthetic-full}, first column), and either \textsc{Query-until-Confident} querying algorithm variant. Both resulting recommendations align with domain intuition.

A key challenge in applying our framework to bite acquisition was calibrating module confidence scores. This involved adapting RT-1 to produce novel confidence scores, and developing a common calibration procedure for all modules that could operate under their varying empirical confidence distributions \ifincludeappendix
    (App.  \ref{sec:calibration-data-confidence-intervals}-\ref{subsec:calibrated-confidence-scores}).
\else
    (App. \appref{sec:calibration-data-confidence-intervals}).
\fi While our confidence estimates were reasonably calibrated, future work could explore conformal prediction \cite{ren2023robots, mullen2024towards} or data-driven calibration methods \cite{karli2025insight}. Future work could also consider module selectors for more complex redundancy structures. While we studied graph-based module selectors for product-of-confidences and sum-of-uncertainties objectives, a hybrid selector could adaptively combine these structures based on which modules are redundant. Finally, longer-term user studies may further differentiate human-in-the-loop algorithms, beyond our observed satisfaction trends. Future evaluations could consider more diverse in-the-wild dishes with richer food sets, more complex geometries, and pre-manipulation skills \cite{jenamani2024flair, ha2024repeat, wu2025savor}.

\begin{acks}
This work was partly funded by NSF CCF 2312774 and NSF OAC-2311521, a LinkedIn Research Award, NSF IIS-244213, NSF IIS \#2132846, CAREER \#2238792, a PCCW Affinito-Stewart Award, and by an AI2050 Early Career Fellowship program at Schmidt Sciences. Research reported in this publication was additionally supported by the Eunice Kennedy Shriver National Institute Of Child Health \& Human Development of the National Institutes of Health and the Office of the Director of the National Institutes of Health under Award Number T32HD113301. The content is solely the responsibility of the authors and does not necessarily represent the official views of the National Institutes of Health. 

The authors would like to additionally thank Yuanchen Ju and Zhanxin Wu for helping with figure creation, and all of the participants in our two in-lab user studies, as well as the two participants in our in-home user study.
\end{acks}

\bibliographystyle{ACM-Reference-Format}
\balance  
\bibliography{sample-base}

\newpage
\clearpage

\ifincludeappendix
    \appendix

\section{Proxy Objectives}
\label{sec:proxy-objectives}

Justification for proxy objective 1: the second term can be a reasonable approximation for $1-\mathbb{E}[r_{\text{task}}]$, where the expectation is over model uncertainties (represented by $c_i$):

\begin{align*}
    \mathbb{E}[r_{\text{task}}] &= P(\cap_i M_i \text{ succeeds}) \\
    &= \prod_{M_i} P(M_i \text{ correct} | \text{predecessor } M_j \text{ correct} ) \\ 
    &= \prod_{M_i \notin M_q} c_i
\end{align*}

Justification for proxy objective 2: Let $u_i=1 - c_i$ be the uncertainty of module $i$. Then we can use union bound to show that:

\begin{align*}
    P(\text{system fail}) &= P(\cup_i M_i \text{ fails}) \\
                          &\leq \sum_i P(M_i \text{ fails}) \\
                          &= \sum_{M_i \notin M_q} u_i + \sum_{M_i \in M_q} 0 \\
                          &= \sum_{M_i \notin M_q} u_i
\end{align*}        

Thus, the proxy objective 2 task component is an upper bound in $P(\text{system fail}) = 1-\mathbb{E}[r_{\text{task}}]$, meaning that proxy objective 2 is a loose upper bound on the original objective, so minimizing proxy objective 2, could also minimize the original objective.

\section{Algorithm Performance Analysis}


Suppose that we have a module graph $G_M$ where we have exactly one module in failure (denoted $M_f$), with all other modules in success. Additionally, we assume that the cost of querying $q_i=0.1$ for all modules. The module in failure has confidence $c_f=0.1$, and all other modules have confidence $c_i=0.1$.

\subsection{GraphQuery}

Recall that in \textsc{GraphQuery}, we assign edge costs $C(e) = 1-c_i$ for the autonomous edges. The graph algorithm thus queries for a module if $1-c_i > \epsilon q$; or:

\begin{equation}
    q < (1-c_i)/\epsilon
\label{eq:test}
\end{equation}

We will never query for any of the modules in success because $c_i=1$ for these modules (as $1-c_i$ will always be 0, and we assume that $q$ is nonnegative, so condition \ref{eq:test} can never be met), so we don't have to worry about querying for modules that are upstream of $M_f$.

In our scenario, we will query for $M_f$ since $q = 0.1$; and $(1-c_i)/\epsilon = (1-0.1) = 0.9$; thus condition \ref{eq:test} is met. Thus, \textsc{GraphQuery} will query correctly at the first timestep.

For fixed $q=0.1$, the critical value of $\epsilon$ above which we would cease to query is $\epsilon_{crit} = (1-c_i)/(q) = (1-0.1)/(0.1) = \boxed{9}$.

For fixed $\epsilon=1$, the critical value of $q$ above which we would cease to query is $q_{crit} = (1-c_i)/(\epsilon) = \boxed{0.9}$

\subsection{MIP}

Recall that we approximate the expected task reward $\mathbb{E}[r_{\text{task}}]$ as follows:

\begin{equation*}
    \mathbb{E}[r_{task}] = \prod_{M_i} P(M_i \text{ correct}) = \prod_{M_i} c_i
\end{equation*}

If we decide to query for a particular module $i$, we assume that $c_i = 1$ since we're using the expert feedback.

\begin{itemize}
    \item Hypothetical $J(\psi_{ms}, \psi_q)$ of not querying for any module: $1-(1^{N-1} \cdot 0.1) = 0.9$
    \item Hypothetical $J(\psi_{ms}, \psi_q)$ of querying for any of the non-failure modules: $q + 1-(1^{N-2} \cdot 0.1 \cdot 1) = q + 0.9 = q + 0.9$
    \item Hypothetical cost of querying for the failure module: $q + 1-(1^{N-1} \cdot 1) = q$ 
\end{itemize}

Thus, we should choose to query for the failure module as long as $q < 0.9$, so in our scenario, MIP will always choose to query for the module in failure at the first timestep.

\section{Bite Acquisition Architecture}
\label{sec:bite-acq-baseline}

Fig.  \ref{fig:method} (left) shows the modular bite acquisition architecture that we use in our work, which is an adaptation of a state-of-the-art architecture \cite{jenamani2024flair}, including  novel foundation model components.

This autonomous system consists of four submodules, each with an associated confidence estimate:
\begin{itemize}

    \item $\mathbf{M_1}$: GPT-4o \cite{hurst2024gpt} food type detector that processes a whole plate RGB image $z_{\text{RGB}} \in \mathbb{R}^{H \times W \times 3}$ and identifies a candidate set of food labels $\mathcal{L} = \{l_1, l_2, \dots, l_K\}$, where $K$ is the number of unique food categories detected (e.g., cherry tomatoes, pineapple, etc.). From this set, $M_1$ selects the single label with the highest confidence score as its output:  

    \[
    M_1(z_{\text{RGB}}) \to l^\ast\]

    \item $\mathbf{M_2}$: GroundingDINO \cite{liu2023grounding} bounding box detector that takes as input the selected food type label $l^\ast$ from $M_1$ together with the plate RGB image $z_{\text{RGB}}$. It outputs one or more bounding boxes $\mathcal{B}(l^\ast) = \{b_1, b_2, \dots, b_J\}$, where each $b_j$ corresponds to a detected instance of the food type $l^\ast$ in the image. Thus,  

    \[
    M_2(z_{\text{RGB}}, l^\ast) \to \mathcal{B}(l^\ast).
    \]

    \item $\mathbf{M_3}$: GPT-4o \cite{hurst2024gpt} skill selector that, given a detected food label $l^\ast$ and its corresponding bounding box $b_i$, predicts the optimal skill $a_i^{h} \in \mathcal{A}_{h}$, where $\mathcal{A}_{h}$ is a set of skills (e.g. Skewering, Scooping, Twirling).

    \[
    M_3(l^\ast, b_i) \to a_i^{h}
    \]

    \item $\mathbf{M_4}$: A VLA model RT-1 that refines the skill action $a_i^{h}$ into precise skill parameters $a_i^l \in \mathbb{R}^4$. It takes in the skill action $a_i^h$ and the cropped image from the corresponding bounding box $b_i$. We adapt RT-1 for modular control by fine-tuning it with a new regression projection head on an aggregated dataset consisting of SPANet samples \cite{SPANet}, along with $\sim$1{,}000 additional labeled images that we collected. 
    This design enables us to attribute RT-1’s uncertainty purely to its skill-parameter prediction, while $M_1$, $M_2$, and $M_3$ handle food classification, location estimation, and skill selection.\\

    \textbf{Skill-specific parameterization.}
    We represent all low-level skills using a unified 2D action vector 
    $a_i^l = (x_1, y_1, x_2, y_2)$, with its semantics based on the selected skill $a_i^h$.
    For \emph{skewering} and \emph{twirling}, the interaction point is defined as the midpoint between the two predicted points,
    \[
    (x_c, y_c) = \left(\frac{x_1 + x_2}{2}, \frac{y_1 + y_2}{2}\right),
    \]
    and the fork tine direction $\theta$ computed as follows:
    \[
    \theta = \arctan(y_2 - y_1, x_2 - x_1).
    \]
    For \emph{scooping}, $(x_1, y_1)$ denotes the start of the scooping motion and $(x_2, y_2)$ denotes the end of the scooping trajectory.
    This parameterization follows prior work on vision-conditioned manipulation primitives (e.g., FLAIR~\cite{jenamani2024flair}), while enabling a unified continuous action space across heterogeneous skills.
    
    \[
    M_4(a_i^{h},b_i) \to a_i^{l}
    \]
    
\end{itemize}

\subsection{Calibration Datasets and Confidence Intervals}
\label{sec:calibration-data-confidence-intervals}
To support uncertainty-aware decision-making in our bite-acquisition pipeline, we adapt the work mentioned in \cite{palempalli2025humanintheloop}. We construct calibration datasets and associated confidence intervals for all modules: $M_1$, $M_2$, $M_3$, $M_4$. Each calibration dataset records per-instance model outputs, their confidence scores, and the corresponding ground-truth labels.

\textbf{Calibration for $M_1$.} We use dishes and all the available plate images from \cite{jenamani2024flair}. For each food type in an image, $M_1$ generates a ranked list of candidate tokens (labels) with log probabilities. These are exponentiated, scaled to percentages, and rounded to two decimal places to yield confidence scores, and the top token (token with the highest confidence score) is selected as the predicted label for that food type.

\begin{itemize}
    \item If the top token matches the ground-truth label for the food type, we add an entry containing the top-5 tokens, their confidence scores, and the ground-truth label to $M_1$'s calibration dataset.
    \item If the ground-truth label is not among the predictions, we substitute a label: either an unmatched food type prediction (if available) or, as a fallback, a matched one from the same image. The substituted token and its associated scores are stored alongside the ground-truth label.  
\end{itemize}

This substitution procedure ensures that every target food type contributes an entry to the calibration dataset, which is necessary for computing reasonable reference intervals. This yields a confidence-based calibration dataset for $M_1$ with 96 samples.

\textbf{Calibration for $M_2$.} For the same set of images used in $M_1$, we iterate over all bounding boxes generated by $M_2$. Each bounding box is manually assessed to determine whether it is \textit{successful} (accurately capturing a target food type) or \textit{unsuccessful} (e.g., enclosing the wrong food type, empty regions, or background noise). For each bounding box, we record the confidence score produced by GroundingDINO. If the bounding box is deemed successful, its confidence score is added to the success list; if unsuccessful, its confidence score is added to the failure list. In this setup, the success scores play the role of the “top-choice” confidence values, while the failure scores serve as the counterpart to the “second-choice” values used in $M_1$ and $M_3$. This yields a calibration dataset of 76 samples for $M_2$ consisting of success (61 samples) and failure (15 samples) confidence distributions that are used to construct intervals analogous to those for the other modules.  

\textbf{Calibration for $M_3$.} The identified (or substituted) label from $M_1$, together with the corresponding bounding box image from $M_2$ of that food type, is passed to $M_3$. To ensure reliable inputs, we restrict this step to bounding boxes deemed valid---those accurately capturing the food type without just empty regions and such. This module outputs candidate skill tokens and their log probabilities, which are processed into percentage confidence scores using the same procedure as $M_1$. Similar to the procedure in $M_1$, the top token is taken as the predicted skill. For each food type and bounding box image pair, we store the top 3 tokens, their confidence scores, and the ground-truth skill label in the calibration dataset for $M_3$. This yields a confidence-based calibration dataset for $M_3$, with 66 samples, analogous to that of $M_1$.  

\textbf{Calibration for $M_4$.} Since RT-1 is deterministic, we introduce stochasticity into its predictions using Monte Carlo Dropout. Specifically, we enable dropout during inference and run the model in batches of 16 forward passes. This produces a distribution over the $M_4$ output space, from which we compute the variance. The variance then serves as our measure of the model’s confidence in its prediction. We used the same bounding boxes as we used in $M_3$, but with the human annotated ground truth label for skills. This yields a confidence-based calibration dataset for $M_4$ with 66 samples.

\textbf{Confidence Intervals.} From each calibration dataset, we compute the mean ($\mu^*$) and standard deviation ($\sigma^*$) of the top tokens' confidence scores. We do the same for the second top tokens, and get $\mu^+$ and $\sigma^+$. Using these values, we define the following confidence intervals:
\[
I_{\text{TopConf}} = [\mu^* - \sigma^*,\ \mu^* + \sigma^*], \quad 
I_{\text{SecondTopConf}} = [\mu^+ - \sigma^+,\ \mu^+ + \sigma^+].
\]

These intervals capture the expected distribution of confidence scores for top and second-ranked predictions across all calibration instances.  

\subsection{Calibrated Confidence Scores}
\label{subsec:calibrated-confidence-scores}

To obtain calibrated confidence scores for all the modules, we apply a rule that maps raw confidence values into a stable binary value. For a given confidence score $c^*$ (corresponding to the top prediction for an instance), we check whether $c^*$ falls outside the calibrated top-token interval or within the second-token interval:  
\[
\text{$Conf_{cal}$}(c^*) = 
\begin{cases}
0, & \text{if } (c^* \notin I_{\text{TopConf}}) \;\; \lor \;\; (c^* \in I_{\text{SecondTopConf}}), \\
1, & \text{otherwise.}
\end{cases}
\]

Here, the function $\text{$Conf_{cal}$}(c^*)$ produces a binary calibrated confidence score: $0$ when the prediction is considered low-confidence (uncertain), and $1$ when the prediction is considered high-confidence (confident). These calibrated scores are later consumed by other components of the pipeline that determine whether or not to query the human for that particular module.  

Grounding the decision in confidence intervals estimated from calibration data allows the system to identify both underconfident correct predictions and overconfident incorrect predictions, producing a more stable confidence signal than raw probabilities alone. By introducing calibrated confidence scores, we ensure that the system reasons over interpretable, statistically grounded intervals rather than noisy probability values, providing more reliable and consistent confidence estimates across all the modules.

\subsection{Retrieval-Augmented Feedback for Error Recovery}
\label{sec:rag-information}
To enable the system to recover from past errors and adapt over time, we integrate a retrieval-augmented generation (RAG) component into both the food type identification module ($M_1$) and the skill selection module ($M_3$). The goal of this component is to incorporate human-provided feedback into a persistent store, allowing the pipeline to retrieve and reuse corrections when similar inputs are encountered in the future. This mechanism provides a means of continual learning from mistakes and reduces repeated queries to the human user.  

\textbf{Embedded Feedback Store.}  
We implement an EmbeddedFeedbackStore that records feedback entries consisting of the plate image (for $m_1$) or bounding box image and food label (for $m_3$), together with the corrected output provided by the human. For $m_1$, embeddings are computed directly from the plate image using the CLIP vision encoder \cite{radford2021learningtransferablevisualmodels}, resulting in a purely visual representation. For $m_3$, embeddings are computed jointly from the bounding box image of the target food item and its textual label using the CLIP vision–language model \cite{radford2021learningtransferablevisualmodels}, capturing both visual and semantic context for skill selection. All embeddings are normalized and stored persistently as vectors alongside the corresponding ground-truth correction, which in our setup refers to the a correct food type label for $M_1$ and the correct skill for $M_3$. So whenever the system queries the human for feedback for a particular module, the corresponding input (image or bounding box image with label) together with its corrected output is added to the feedback store for future retrieval and reuse.   

\textbf{Retrieval.}  
At inference time, when $M_1$ or $M_3$ produces a prediction, the feedback store computes the CLIP embedding for the current input and retrieves the most similar past entry using cosine similarity. If the best match exceeds a similarity threshold, the stored correction is reused. This correction is converted into a probability distribution: the retrieved food type label (or skill) is assigned a probability proportional to the similarity score, and the remaining probability mass is evenly distributed across the other candidate tokens. If no sufficiently similar correction is found, the pipeline defaults to the module’s prediction.  

This RAG-based mechanism allows the pipeline to learn incrementally from its mistakes. By leveraging similarity search, the system avoids repeating prior errors on similar inputs and reduces unnecessary human queries. This provides a lightweight but effective form of continual adaptation. 

\subsection{Query questions and feedback in user study}
\label{sec:bite-acq-queries}
Four types of questions can be asked when executing the \textit{Always Query} method and our selected method in the user study:
\begin{enumerate}
    \item The robot will ask the user for help by asking the following question: “Could you tell me a food item that is on the plate?” Users can respond by providing any valid food item that is present on the plate. 
    \item The robot will ask the user for help by displaying the plate image on the tablet. Users can respond by tapping 2 opposite corners of the box on the screen to create a bounding box. 
    \item The robot will ask the user for help by asking you the following question: “For the food item on the plate, what skill should I use?”. Users can respond by telling the robot the skill that it should use. 
    \item The robot will ask the user for help by displaying the bounding box image on the tablet. Users can respond by tapping 2 points in the following manner: for skewering, tap two points that define the longer edge of the food item; for scooping, tap the start and the end of the scoop; for twirling, tap two points around where the user thinks the fork should twirl the food item. 
\end{enumerate}

\section{Bite Acquisition Architecture with Module Heterogeneity}
\label{sec:bite-acq-hetero}

This architecture is similar to that described in Sec. \ref{sec:bite-acq-baseline}, with the following modifications:

\begin{itemize}
\item $\mathbf{M_1}$: The food detector is faulty, producing a random food label with probability 0.9 (which lies outside of the set of food items encountered in the "savory salad" and "Thanksgiving meal" plates). In this scenario, the food detector assigns a confidence of $c_1 = 0.65$. In the event of a success, $M_1$ produces the label with the highest confidence score $\ell^*$ as before, but with a confidence score $c_1 = \frac{c^*}{c_{max, M_1}}$, where $c^*$ is the raw confidence score (Sec. \ref{subsec:calibrated-confidence-scores}) and $c_{max}$ is the maximum raw confidence score observed in the $M_1$ calibration set.
\item $\mathbf{M_2}$: Similar to Sec. \ref{sec:bite-acq-baseline}, but with a confidence score $c_2 = \frac{c^*}{c_{max}}$, where $c^*$ is the raw confidence score (Sec. \ref{subsec:calibrated-confidence-scores}) and $c_{max, M_2}$ is the maximum raw confidence score observed in the $M_2$ calibration set.
\item $\mathbf{M_3}$: Similar to Sec. \ref{sec:bite-acq-baseline}, but with a confidence score $c_3 = \frac{c^*}{c_{max}}$, where $c^*$ is the raw confidence score (Sec. \ref{subsec:calibrated-confidence-scores}) and $c_{max, M_3}$ is the maximum raw confidence score observed in the $M_3$ calibration set.
\item $\mathbf{M_4}$: Unchanged compared to Sec. \ref{sec:bite-acq-baseline}.
\end{itemize}

\section{Workload Model Conditioning}

To predict workload for each of the bite acquisition modules (App.  \ref{sec:bite-acq-baseline}), we set the query type variables for the predictive workload model as follows \cite{banerjee2024ask}: 

\begin{itemize}
    \item $M_1$: $d_t$ = easy, $resp_t$ = MCQ, $dist_t$ = “no distraction task”
    \item $M_2$: $d_t$ = easy, $resp_t$ = BB, $dist_t$ = “no distraction task”
    \item $M_3$: $d_t$ = easy, $resp_t$ = MCQ, $dist_t$ = “no distraction task”
    \item $M_4$: $d_t$ = hard, $resp_t$ = BB, $dist_t$ = “no distraction task”
\end{itemize}

\label{app:workload-model}

\section{Additional Synthetic Experiments: Systematic Ablations}
\label{app:synthetic-exps}

\subsection{Additional experimental setup details.}
\label{app:synthetic-exp-setup}

We set the GraphQuery hyperparameter $\epsilon = 1$, and we assume equal weight between querying and task reward ($w = 0.5$).

Independent variable settings considered:
\begin{itemize}
    \item Number of modules $N$: [3, 5, 10, 15, 18, 25, 50, 75, 100]
    \item Graph redundancy structure $G_M$: [fully redundant (all-OR), fully dependent (all-AND), fully redundant followed by fully dependent (OR-then-AND), fully dependent followed by fully redundant (AND-then-OR)]
    \item Module confidences $c_i$: [(1.0, 0.1), (0.9, 0.2), (0.8, 0.3), (0.7, 0.4)], where ($c_h$,$c_l$) are the high confidence value and low confidence value, respectively, described in Sec. \ref{sec:synthetic-results}. We assume that 3 of the modules in the module graph $G_M$ are assigned the low confidence value, and the rest are assigned the high confidence value.
    \item Query costs $q_i$: [0.08, 0.16, 0.32, 0.64]
\end{itemize}

Below, we vary each independent variable in isolation, fixing the other variables to the setting described in Sec. \ref{sec:synthetic-results}.

\subsection{Number of modules $N$}

\begin{figure*}[!ht]
    \centering
    \includegraphics[width=\linewidth,alt={Multi-panel plots showing the effect of increasing the number of modules on query cost, failed attempts, computation time, task cost outcomes, and total timesteps. As the number of modules increases, confidence-based and workload-aware querying maintain low failure rates and incorrect outcomes with moderate computation time.}]{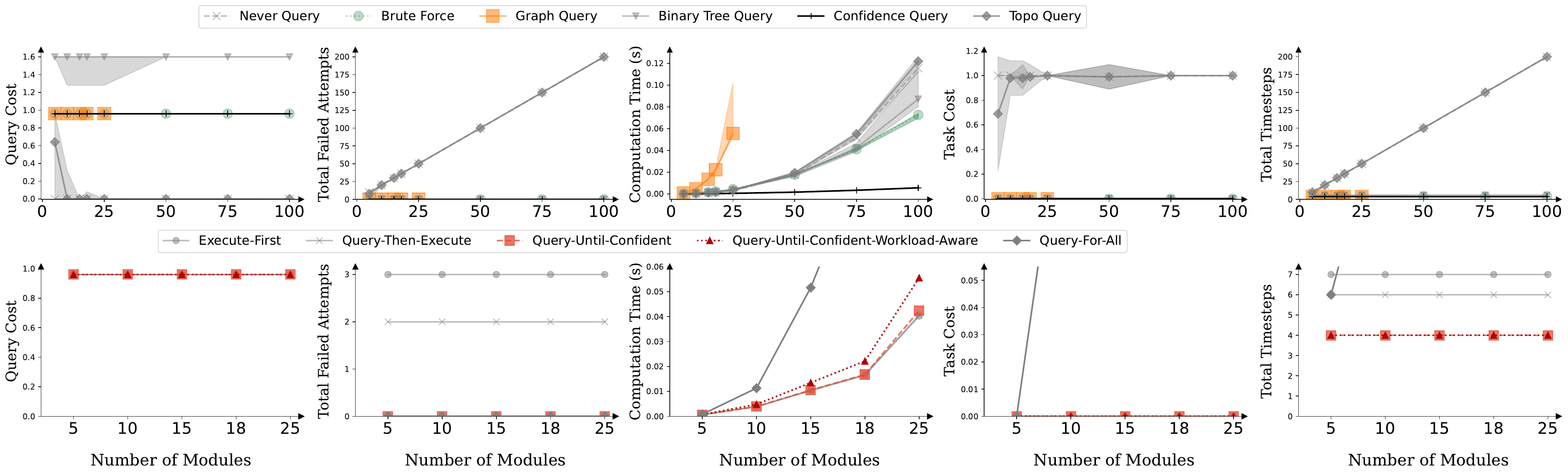}
    \caption{(top) Module selector comparison for varying the number of modules $N$ in the module graph $G_M$, for the \textsc{Query-Until-Confident-Workload-Aware} querying algorithm. (bottom) Querying algorithm comparison for varying the number of modules $N$ in the module graph $G_M$, for the \textsc{GraphQuery} module selector. Plots show median, upper quartile, and lower quartile values across 100 trials (mean for Task Cost).}
    \Description{Multi-panel plots showing the effect of increasing the number of modules on query cost, failed attempts, computation time, task cost outcomes, and total timesteps. As the number of modules increases, confidence-based and workload-aware querying maintain low failure rates and incorrect outcomes with moderate computation time.}
    \label{fig:num-modules}
\end{figure*}

Full results for varying the number of modules $N$ across module selectors and querying algorithms are shown in Fig. \ref{fig:num-modules}. Key takeaways:

\begin{itemize}
    \item No metrics (besides \textit{Computation Time}) vary as a function of number of modules, except in different variable settings (e.g. closer confidence values) where numerical stability is a concern.
    \item When confidence values differ from $(1, 0.1)$, module selector performance degrades at larger graph sizes due to numerical underflow. We find that \textsc{BinaryTreeQuery} is the most sensitive, followed by \textsc{GraphQuery}, followed by \textsc{BruteForce}.
\end{itemize}

\textit{Recommendations.} For module selectors, we recommend using either \textsc{BruteForce}, \textsc{GraphQuery}, or \textsc{ConfidenceQuery}, with the latter being the most scalable to increasing module graph size. For querying algorithms, we recommend \textsc{Query-until-Confident} or \textsc{Query-until-Confident-Workload-Aware} as they minimize \textit{Total Failed Attempts} and \textit{Total Timesteps}.

\subsection{Graph structure $G_M$}

Full results for varying the graph structure $G_M$ across module selectors and querying algorithms are shown in Fig. \ref{fig:redundancy}. Key takeaways:

\subsubsection{Module selectors.}

\begin{itemize}
    \item \textit{Query Cost.} For the variable setting shown in Fig. \ref{fig:redundancy}, we find that \textit{Query Cost} is insensitive to the redundancy structure. Across other variable settings, we generally observe that the query cost is lower for the all-OR redundancy structure, compared to the other redundancy structures, because fewer queries are needed for the overall system output to be correct. Additionally, the query cost for the OR-then-AND structure is slightly higher than that for the AND-then-OR redundancy structure (because overall system success is more likely when the final module is an OR module). 
    \item \textit{Total Failed Attempts.} For the variable setting shown in Fig. \ref{fig:redundancy}, we find that \textit{Total Failed Attempts} is insensitive to the redundancy structure (except for \textsc{NeverQuery}). Across other variable settings, we observe that \textit{Total Failed Attempts} is lower for the all-OR redundancy structure. We additionally observe that the relative ordering of module selectors does not generally depend on the redundancy structure, and proceeds roughly as follows (in descending order): \textsc{NeverQuery}, \textsc{BinaryTreeQuery}, \textsc{BruteForce}, followed by \textsc{GraphQuery}. 
    \item \textit{Computation Time.} We observe that \textit{Computation Time} is lower for the all-OR redundancy structure. \textsc{GraphQuery} tends to have a higher \textit{Computation Time} than the other module selectors (regardless of the redundancy structure) due to the computational complexity of parallel graph creation + parallel shortest-path searches.
    \item \textit{Task Cost.} For the variable setting shown in Fig. \ref{fig:redundancy}, we find that \textit{Task Cost} is insensitive to the redundancy structure (except for \textsc{NeverQuery}). In other settings, \textsc{NeverQuery} and \textsc{BinaryTreeQuery} only achieve complete success for the all-OR architecture, and have no success for the other redundancies.
    \item \textit{Total Timesteps.} For the variable setting shown in Fig. \ref{fig:redundancy}, we find that \textit{Total Timesteps} is insensitive to the redundancy structure (except for \textsc{NeverQuery}), with \textsc{BinaryTreeQuery} having a slightly higher value across-the-board compared to the other module selectors.
\end{itemize}

\subsubsection{Querying algorithms.}

\begin{itemize}
    \item \textit{Query Cost.} For the variable setting shown in Fig. \ref{fig:redundancy}, generally lower for most querying algorithms in all-OR redundancy structure, compared to other structures. In other low query cost settings: (1) higher for \textsc{Query-until-Confident-Workload-Aware} compared to other methods, (2) higher in the hybrid structures (compared to the pure structures) across-the-board.
    \item \textit{Total Failed Attempts.} Lower for all querying algorithms in all-OR redundancy structure, compared to other structures. We note the following rough ordering across querying algorithms, regardless of graph structure, in descending order: \textsc{Execute-First} > \textsc{Query-then-Execute} > \textsc{Query-until-Confident} $\sim$ \textsc{Query-until-Confident-Workload-Aware}. We also notice the following ordering across graph structures, regardless of querying algorithm, in descending order: OR-then-AND > all-AND > AND-then-OR > all-OR.
    \item \textit{Computation Time.} Lower for all querying algorithms in all-OR redundancy structure, compared to other structures. We note that \textsc{Query-Until-Confident-Workload-Aware} has slightly higher computation time than other querying algorithms, independent of graph structure.
    \item \textit{Task Cost.} Fairly consistent across graph structures. In other settings, \textsc{Query-Until-Confident-Workload-Aware} tends to have the highest success rate across all querying algorithms, as it is able to recover in some settings where other querying algorithms fail (e.g. AND-then-OR, and OR-then-AND structures).
    \item \textit{Total Timesteps.} Lower for most querying algorithms in all-OR redundancy structure, compared to other structures. WE note the following roughly consistent ordering across querying algorithms, regardless of graph structure, in descending order: \textsc{Execute-First} > \textsc{Query-then-Execute} > \textsc{Query-until-Confident} $\sim$ \textsc{Query-until-Confident-Workload-Aware}, consistent with the \textit{Total Failed Attempts} trend.
\end{itemize}

\textit{Recommendations.} For module selectors, we recommend using either \textsc{BruteForce}, \textsc{GraphQuery}, or \textsc{ConfidenceQuery}, as all 3 have the best performance across redundancy structures. For querying algorithms, we recommend \textsc{Query-until-Confident} or \textsc{Query-until-Confident-Workload-Aware} perform best across the metrics (for all redundancies except all-OR, where \textsc{Execute-First} and \textsc{Query-and-Execute} outperform the other querying algorithms in \textit{Query Cost} and \textit{Total Timesteps}).

\begin{figure*}[!ht]
    \centering
    \includegraphics[width=\linewidth,alt={Multi-panel plots comparing module selectors and querying algorithms across different module graph redundancy structures, including all-AND, all-OR, AND-then-OR, and OR-then-AND. Confidence-based and workload-aware querying strategies (both module selectors and querying algorithms) consistently reduce failed attempts and incorrect outcomes across dependency types.}]{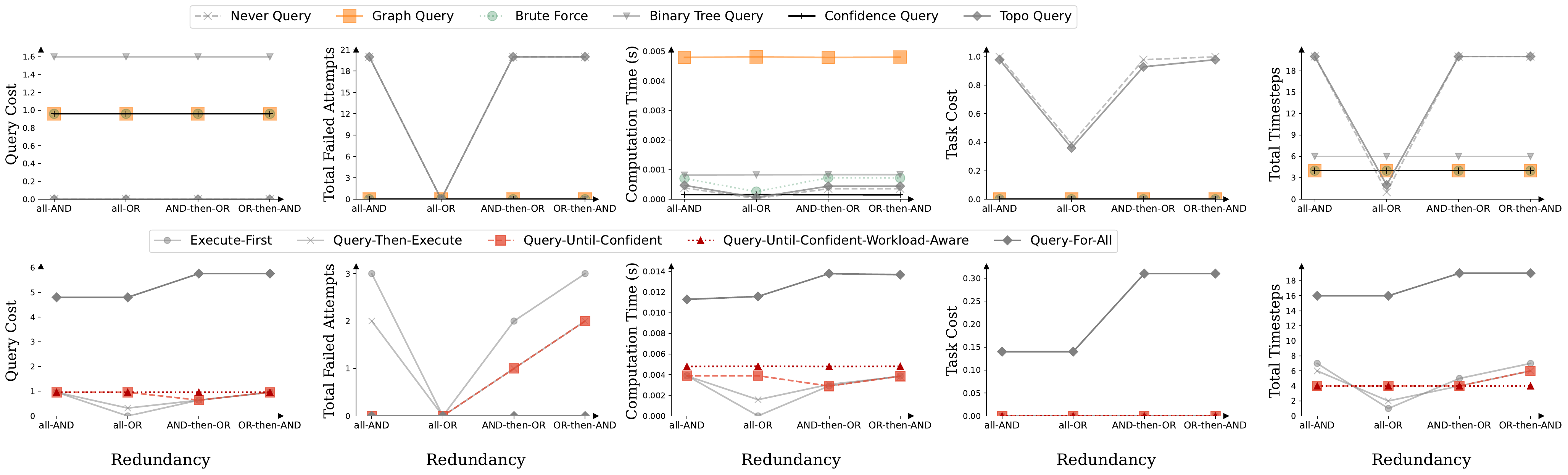}
    \caption{Varying the graph redundancy structure for fixed querying algorithm (\textsc{Query-until-confident-workload-aware}; top) and fixed module selector (\textsc{GraphQuery}; bottom). Plots show median values across 100 trials (mean for Task Cost).}
    \Description{Multi-panel plots comparing module selectors and querying algorithms across different module graph redundancy structures, including all-AND, all-OR, AND-then-OR, and OR-then-AND. Confidence-based and workload-aware querying strategies (both module selectors and querying algorithms) consistently reduce failed attempts and incorrect outcomes across dependency types.}
    \label{fig:redundancy}
\end{figure*}

\subsection{Confidence values $c_i$}

Full results for varying the confidence values $c_i$ across module selectors and querying algorithms are shown in Fig. \ref{fig:confidence}. Key takeaways:

\begin{figure*}[!ht]
    \centering
    \includegraphics[width=\linewidth,alt={Multi-panel plots evaluating query strategies under varying module confidence levels. As confidences become less separated, all module selectors and querying algorithms degrade in performance, with the confidence- and workload-aware module selectors being the most robust, and the Query-Until-Confident querying algorithm being the most robust.}]{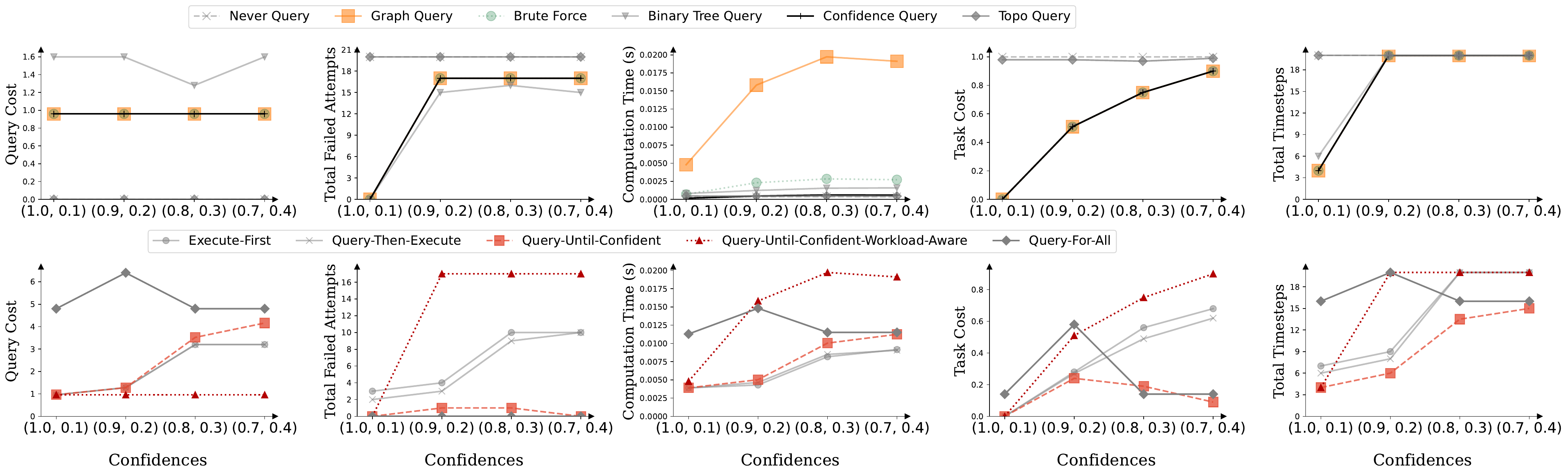}
    \caption{Varying the confidence levels for fixed querying algorithm (\textsc{Query-until-confident-workload-aware}; top) and fixed module selector (\textsc{GraphQuery}; bottom). Plots show median values across 100 trials (mean for Task Cost).}
        \Description{Multi-panel plots evaluating query strategies under varying module confidence levels. As confidences become less separated, all module selectors and querying algorithms degrade in performance, with the confidence- and workload-aware module selectors being the most robust, and the Query-Until-Confident querying algorithm being the most robust.}
    \label{fig:confidence}
\end{figure*}

\subsubsection{Module selectors.}

\begin{itemize}
    \item \textit{Query Cost.} Generally higher when confidences are closer together. Regardless of confidence level, generally follows the following pattern in descending order: \textsc{BinaryTreeQuery} > \textsc{GraphQuery} $\sim$ \textsc{BruteForce} > \textsc{Never Query}.
    \item \textit{Total Failed Attempts.} Generally higher when confidences are closer together. Regardless of confidence level, generally follows the following pattern in descending order: \textsc{NeverQuery}, \textsc{GraphQuery} $\sim$ \textsc{BruteForce}, \textsc{BinaryTreeQuery}. 
    \item \textit{Computation Time.} Generally higher when confidences are closer together. \textsc{GraphQuery} has much higher Computation Time, followed by \textsc{BruteForce}, \textsc{BinaryTreeQuery} and \textsc{NeverQuery}.
    \item \textit{Task Cost.} All module selectors (except \textsc{NeverQuery}) tend to increase from 0.0 to 1.0 with less-separated confidences. \textsc{NeverQuery} always has a value of $1.0$, except in the all-OR setting (not shown in Fig. \ref{fig:confidence}).
    \item \textit{Total Timesteps.} Generally higher when confidences are closer together (for all module selectors except \textsc{NeverQuery}). 
\end{itemize}

\subsubsection{Querying algorithms.}

\begin{itemize}
    \item \textit{Query Cost.} Generally higher when confidences are closer together. For low query cost settings (including Fig. \ref{fig:confidence}), \textsc{Query-until-Confident-Workload-Aware} has the highest \textit{Query Cost} compared to other querying algorithms.
    \item \textit{Total Failed Attempts.} Generally higher across all querying algorithms when confidences are closer together. Lower for \textsc{Query-until-Confident} than \textsc{Execute-First} and \textsc{Query-then-Execute}.
    \item \textit{Computation Time.} Generally higher when confidences are closer together. Slightly higher for \textsc{Query-until-Confident-Workload-Aware} compared to the other querying algorithms.
    \item \textit{Task Cost.} Generally higher when confidences are closer together. Lower for \textsc{Query-until-Confident} than \textsc{Execute-First} and \textsc{Query-then-Execute}.
    \item \textit{Total Timesteps.} Generally higher when confidences are closer together. Regardless of confidence score, \textit{Total Timesteps} is generally lower for \textsc{Query-until-Confident} compared to \textsc{Execute-First} and \textsc{Query-then-Execute}.
\end{itemize}

\textit{Recommendations.} For module selectors, we recommend using either \textsc{BruteForce} or \textsc{GraphQuery}, as they are the most competitive across metrics regardless of the confidence level. For querying algorithms, we recommend \textsc{Query-until-Confident} (unless in the all-OR setting, in which case \textsc{Execute-First} or \textsc{Query-then-Execute} are the best at minimizing \textit{Total Timesteps}).

\subsection{Query Cost $q_i$}

Full results for varying the confidence values $c_i$ across module selectors and querying algorithms are shown in Fig. \ref{fig:querycost}. Key takeaways:

\begin{figure*}[!ht]
    \centering
    \includegraphics[width=\linewidth,alt={Plots showing the impact of increasing query cost on module selector and querying algorithm performance. Performance of all module selectors and querying algorithms (except for total query cost) is relatively invariant to the query cost level, with the query-until-confident querying algorithms having the lowest total timesteps across all query cost levels.}]{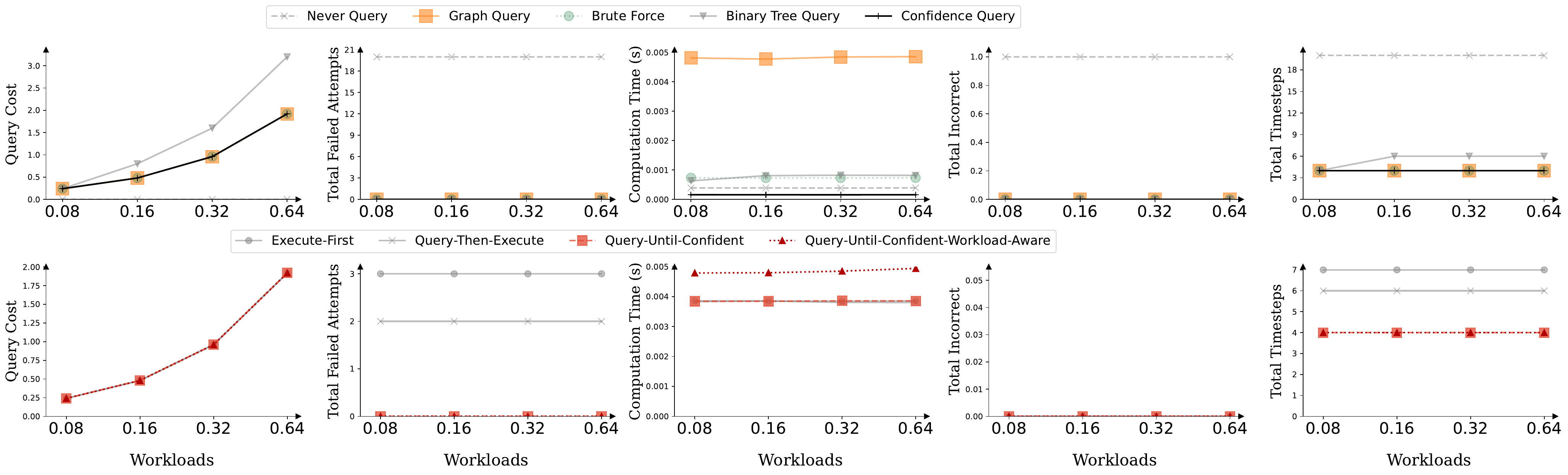}
    \caption{Varying the module query costs for fixed querying algorithm (\textsc{Query-until-confident-workload-aware}; top) and fixed module selector (\textsc{GraphQuery}; bottom). Plots show median values across 100 trials (mean for Task Cost).}
    \Description{Plots showing the impact of increasing query cost on module selector and querying algorithm performance. Performance of all module selectors and querying algorithms (except for total query cost) is relatively invariant to the query cost level, with the query-until-confident querying algorithms having the lowest total timesteps across all query cost levels.}
    \label{fig:querycost}
\end{figure*}

\subsubsection{Module selectors.}

\begin{itemize}
    \item \textit{Query Cost.} Higher \textit{Query Cost} across-the-board as we increase the cost of querying, which is expected. \textsc{BinaryTreeQuery} generally has the highest \textit{Query Cost} (regardless of module query cost), while \textsc{BruteForce} and \textsc{GraphQuery} are lower.
    \item \textit{Total Failed Attempts.} Relatively invariant to the query cost level. 
    \item \textit{Computation Time.} Generally highest for \textsc{GraphQuery}, regardless of the query cost level. In ascending order for the other module selectors, generally see \textsc{NeverQuery}, then \textsc{BruteForce}, then \textsc{BinaryTreeQuery}.
    \item \textit{Task Cost.} Generally consistent for all methods, regardless of query cost level. For higher query cost settings and confidence settings with less separation (not shown in Fig. \ref{fig:querycost}), we do observe consistent a \textit{Task Cost} value of 0.0 across all settings for \textsc{BinaryTreeQuery} and \textsc{NeverQuery}.
    \item \textit{Total Timesteps.} Values are also generally invariant as a function of query cost, with \textsc{BinaryTreeQuery} having a slightly higher value compared to \textsc{BruteForce} and \textsc{GraphQuery}.
\end{itemize}

\subsubsection{Querying algorithms.}

{
\sloppy
\begin{itemize}
    \item \textit{Query Cost.} \textit{Query Cost} generally increases as query cost increases, as expected.
    \item \textit{Total Failed Attempts.} Typical trend is \textsc{Execute-First} > \textsc{Query-then-Execute} > \textsc{Query-until-Confident} $\sim$ \textsc{Query-until-Confident-Workload-Aware}, regardless of the query cost setting. In some settings, \textit{Total Failed Attempts} increases slightly for \textsc{Query-until-Confident-\allowbreak Workload-Aware} as a function of query cost.
    \item \textit{Computation Time.} Generally comparable across querying algorithms, invariant to query cost setting. In some scenarios (e.g. in Fig. \ref{fig:querycost}), \textsc{Query-until-Confident-Workload-Aware} has a higher computation time than the other querying algorithms.
    \item \textit{Task Cost.} Generally comparable across querying algorithms (0.0), invariant to query cost setting. Sometimes observe degradation to 1.0 (in module settings with smaller separation in confidences, not shown in Fig. \ref{fig:querycost}) for all methods besides \textsc{Query-until-Confident-Workload-Aware}.
    \item \textit{Total Timesteps.} Generally invariant to query cost setting. Regardless of confidence score, \textit{Total Timesteps} is generally lower for \textsc{Query-until-Confident-Workload-Aware} and \textsc{Query-until-Confident} compared to \textsc{Execute-First} and \textsc{Query-then-Execute} (except for the all-OR redundancy structure, where the trend is flipped).
\end{itemize}
}

\textit{Recommendations.} For module selectors, we recommend using either \textsc{BruteForce}, \textsc{GraphQuery}, or \textsc{ConfidenceQuery} as they are the most competitive across metrics regardless of the module query cost. For querying algorithms, we recommend \textsc{Query-until-Confident-Workload-Aware} or \textsc{Query-until-Confident} (\textsc{Query-until-Confident-\allowbreak Workload-Aware} if minimizing \textit{Query Cost} and \textit{Total Failed Attempts} is most important; \textsc{Query-until-Confident} if minimizing \textit{Computation Time} is the most important).

\subsection{Additional Ablation: Expert Confidence}
\label{appendix:noisy-expert}

\begin{figure*}[!ht]
    \centering
    \includegraphics[width=\linewidth,alt={Multi-panel plots examining the effect of expert confidence on module selector and querying algorithm behavior and performance. As expert confidence decreases, performance of all module selectors and querying algortihms degrades. Query-until-confident querying algorithm variants perform best in high expert confidence regimes, while execute-first and query-then-execute perform best in low expert confidence regimes.}]{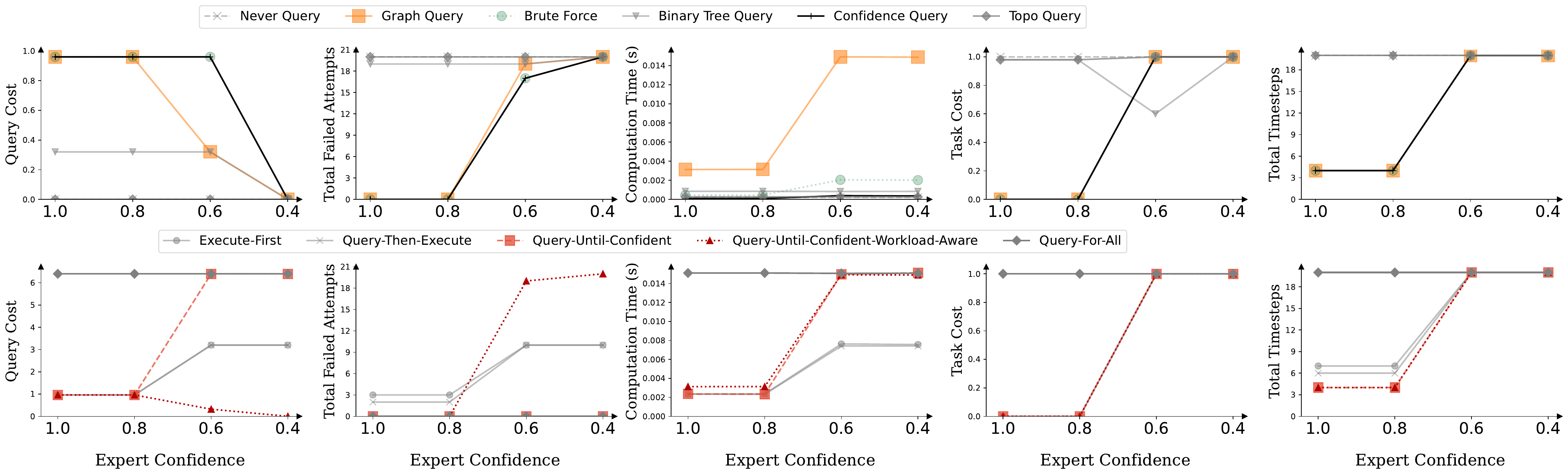}
    \caption{Varying the expert query confidence for fixed querying algorithm (\textsc{Query-until-Confident-Workload-Aware}; top) and fixed module selector (\textsc{Graph-Query}; bottom). Plots show median values across 100 trials (mean for Task Cost).}
    \Description{Multi-panel plots examining the effect of expert confidence on module selector and querying algorithm behavior and performance. As expert confidence decreases, performance of all module selectors and querying algortihms degrades. Query-until-confident querying algorithm variants perform best in high expert confidence regimes, while execute-first and query-then-execute perform best in low expert confidence regimes.}
    \label{fig:noisyexpert}
\end{figure*}

We consider an ablation over an additional user variable: the expert confidence value ($c_{\text{expert}}$, introduced in Sec. \ref{sec:problem-formulation}), to understand how imperfect human feedback affects the performance of the module selectors and querying algorithms. In our simulation, the expert confidence $c_{\text{expert}}$ represents the probability that a module query produces the correct logical value for that module. We consider the following settings for $c_{\text{expert}}$: $[1, 0.8, 0.6, 0.4]$, which range from a perfect expert (considered in Sec. \ref{sec:synthetic-results}) to a less confident expert. Additionally, we adapt the module selectors to assign a confidence of $c_{\text{expert}}$ to modules that have already been queried.

\subsubsection{Module selectors.}

We find that performance degrades for all module selectors as $c_{\text{expert}}$ decreases (Fig. \ref{fig:noisyexpert}), with \textit{Task Cost} degrading to 0 at the lowest confidence setting ($c_{\text{expert}}=0.4$). For sufficiently low expert confidence, the module selectors cannot determine whether a module has received the correct feedback, leading to redundant querying of the same (upstream) module. Across the module selectors, we find that \textsc{Brute Force}, \textsc{GraphQuery}, and \textsc{ConfidenceQuery} perform best in the high expert confidence regime. We also find that \textsc{BinaryTreeQuery} tends to under-query regardless of expert confidence value, leading to high \textit{Task Cost}.

\subsubsection{Querying algorithms.}

We again find that all querying algorithms degrade in performance as $c_{\text{expert}}$ decreases (Fig. \ref{fig:noisyexpert}). For higher expert confidence values, 
we find that the two \textsc{Query-Until-Confident} variants are the best in \textit{Total Timesteps} and \textit{Total Failed Attempts}. For lower expert confidence values, we note that \textsc{Query-Until-Confident} tends to over-query, because the proxy task objective never becomes high enough to stop querying, and \textsc{Query-Until-Confident-Workload-Aware} tends to under-query, as the confidence gain due to querying is insufficient to overcome the query cost.

\textit{Recommendations.} For intermediate-to-high expert confidence values, we would recommend either the \textsc{BruteForce}, \textsc{GraphQuery}, or \textsc{ConfidenceQuery} module selectors, as they achieve the lowest \textit{Task Cost} with low \textit{Total Timesteps}. Additionally, we would recommend the two \textsc{Query-Until-Confident} querying algorithms if minimizing \textit{Total Timesteps} is the most important, and the \textsc{Execute-First} and \textsc{Query-then-Execute} querying algorithms in other scenarios.

\section{Additional Synthetic Experiments: Module Heterogeneity}
\label{appendix:cq-gq-comprison}

Fig.  \ref{fig:gq-cq-comparison} compares the \textsc{GraphQuery} vs \textsc{ConfidenceQuery} module selectors for additional values of the module confidences and query cost variance $\beta$ (with fixed querying algorithm \textsc{Query-Until-Confident-Workload-Aware}). We find that \textsc{GraphQuery} performs at least as well as \textsc{ConfidenceQuery} across these experimental settings.

\begin{figure*}[!ht]
    \centering
    \includegraphics[width=\linewidth,alt={Plots comparing graph-based module selection and confidence-based module selection under increasing module heterogeneity and variance. Graph-based module selection consistently achieves fewer incorrect outcomes and failed attempts across variance levels, particularly in higher-variance settings, compared to confidence-based module selection.}]{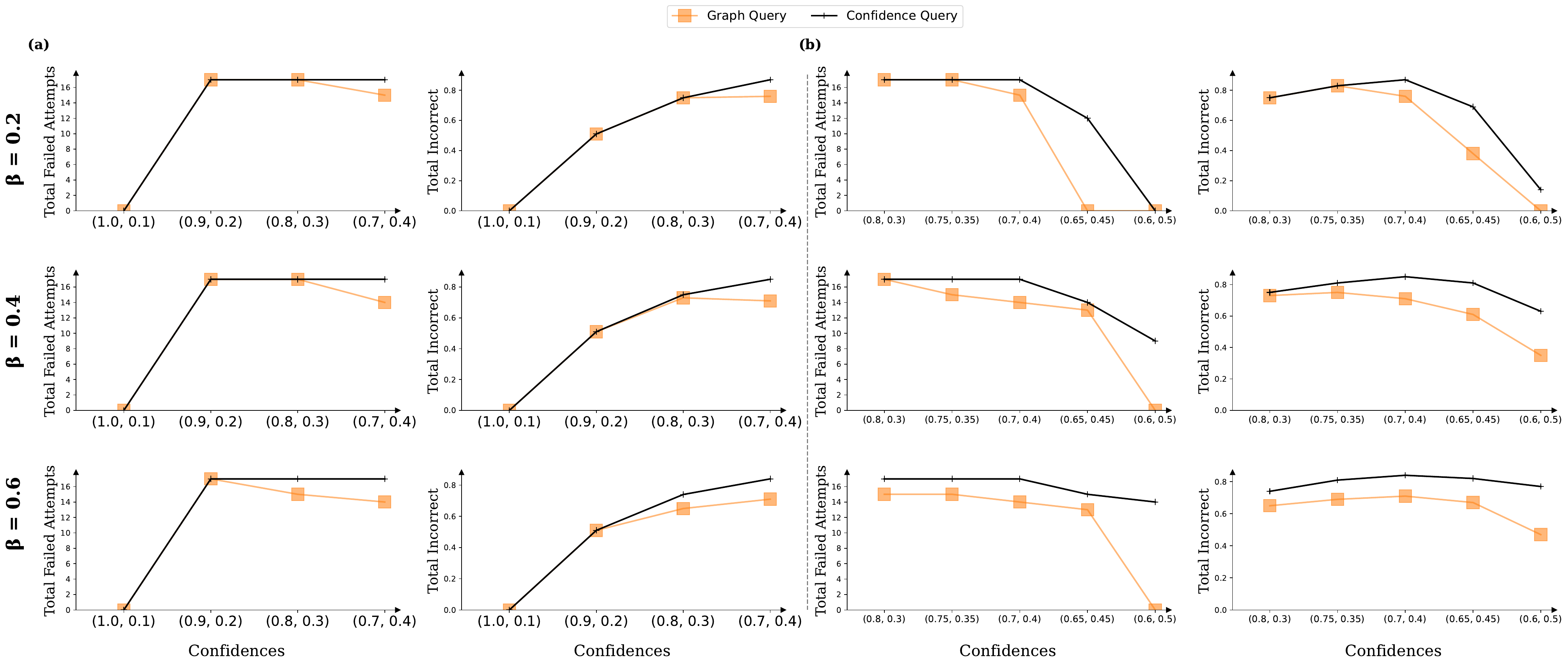}
    \caption{Comparison between \textsc{GraphQuery} and \textsc{ConfidenceQuery}, as a function of varying module confidences and query cost uniform noise $\beta$ (for fixed querying algorithm: \textsc{Query-Until-Confident-Workload-Aware}), both in less overlapping (left) and more overlapping (right) confidence regimes. Plots show median values across 100 trials (mean for Task Cost).}
        \Description{Plots comparing graph-based module selection and confidence-based module selection under increasing module heterogeneity and variance. Graph-based module selection consistently achieves fewer incorrect outcomes and failed attempts across variance levels, particularly in higher-variance settings, compared to confidence-based module selection.}
    \label{fig:gq-cq-comparison}
\end{figure*}

\section{User Study Metrics}
\label{sec:user-study-metrics}
\subsection{Subjective metrics} We asked the participants the following questions in terms of Mental/Physical Demand, Effort, Subjective Success, and Satisfaction, all on a Likert scale from 1-5:
\begin{enumerate}
    \item \textbf{Mental/Physical Demand}: For the last method, how mentally/physically demanding was it for the robot to query you? 
    \item \textbf{Effort}: For the last method, how hard did you have to work to make the robot pick up food items?
    \item \textbf{Subjective Success}: For the last method, how successful was the robot in picking up food items? 
    \item \textbf{Satisfaction}: For the last method, how satisfied are you with how the robot balanced between trying to pickup independently when possible and asking for help when required?
\end{enumerate}

\subsection{Objective metrics} We define the following 3 objective metrics:
\begin{enumerate}
    \item \textbf{Mean Queries Per Plate}: $\frac{1}{B} \sum_{b=1}^B \sum_{t=1}^{T_b} \mathbf{1}$\{\text{queried at }$t\}$, where $B$ is the number of bites per plate and $T_b$ is the number of timesteps needed for bite $b$. 
    \item \textbf{Mean Executions Per Plate}: $\frac{1}{B} \sum_{b=1}^B \sum_{t=1}^{T_b} \mathbf{1}$\{\text{executed at }$t\}$, where $B$ and $T_b$ are defined above.
    \item \textbf{Mean Successful Bites Per Plate}:  $\frac{1}{B} \sum_{b=1}^B \mathbf{1}$\{b \text{succeeded}$\}$, where $B$ is defined above and we define bite $b$ to have succeeded if the robot acquired the bite after $T_b$ timesteps.
\end{enumerate}
\fi

\end{document}